\pdfoutput=1 % arXiv: force pdfLaTeX processing
\documentclass[10pt,twocolumn,letterpaper]{article}

\usepackage{cvpr}              % To produce the CAMERA-READY version
\definecolor{cvprblue}{rgb}{0.21,0.49,0.74}
\usepackage[pagebackref,breaklinks,colorlinks,allcolors=cvprblue]{hyperref}
\usepackage[utf8]{inputenc}
\usepackage{amsmath,amsfonts,amssymb}
\usepackage{array,booktabs,colortbl,multirow,tabularx}
\usepackage{algorithm,algpseudocode}
\usepackage{enumitem}
\usepackage{longtable}
\usepackage{array}
\usepackage{listings}
\usepackage[most]{tcolorbox}
\usepackage{newfloat}
\usepackage{makecell}
\usepackage{graphicx}
\usepackage{soul}
\usepackage{flafter}
\usepackage{float}
\definecolor{rankone}{gray}{0.85}
\definecolor{ranktwo}{gray}{0.93}
\usepackage{xcolor}
\definecolor{SoftBlue}{RGB}{230, 240, 255}
\definecolor{NiceGreen}{RGB}{2,158,115} 
\definecolor{NiceRed}{RGB}{213,94,0}

\definecolor{headerblue}{RGB}{240, 248, 255}
\sethlcolor{headerblue}
\newcommand{\myfinding}[1]{%
  \vspace{0.3em}\noindent\hl{\textbf{#1.}}}

\usepackage{pifont} 
\newtcolorbox{finding}[1]{%
  breakable, enhanced,
  colback=black!5, colframe=black!45,
  boxrule=0.6pt, arc=4pt,
  left=8pt, right=8pt, top=6pt, bottom=6pt,
  before skip=8pt, after skip=8pt,
  before upper={\bfseries\itshape Finding~#1:\quad\normalfont}%
}
\usepackage{amsthm}
\theoremstyle{definition}

\newcolumntype{L}[1]{>{\raggedright\arraybackslash}p{#1}}
\newcolumntype{M}[1]{>{\raggedright\arraybackslash}m{#1}}
\newcolumntype{B}[1]{>{\bfseries\raggedright\arraybackslash}m{#1}}

  \DeclareRobustCommand{\apicode}[1]{\begingroup\Urlmuskip=0mu plus 1mu\urlstyle{tt}\nolinkurl{#1}\endgroup}
\newcommand{\best}[1]{\textbf{#1}}                    % 最优：加粗
\newcommand{\second}[1]{\cellcolor{SoftBlue!40}#1}    % 次优：浅蓝底

\definecolor{asActive}{HTML}{FFF7E8}   % Active 列底色（暖色，避免和 \second 的 SoftBlue 混淆）
\definecolor{asGroup}{HTML}{F2F3F5}    % 模型分组标题行底色
\definecolor{asMuted}{HTML}{8A8F98}    % 占位符与题型小字
\newcommand{\na}{\textcolor{asMuted}{--}}                                                      % 暂无数据
\newcommand{\gain}[1]{\ifdim #1pt<0pt\textcolor{NiceRed}{$#1$}\else\textcolor{NiceGreen}{$+#1$}\fi} % 探索收益，自动上色
  \newcommand{\cmark}{\ding{51}}
  \newcommand{\xmark}{\ding{55}}

\usepackage{tikz}  
\usetikzlibrary{arrows.meta} 
\usepackage{fontawesome5}   % 提供 \faGlobe 和 \faGithub，pdfLaTeX 可以直接用
\newcommand{\hficon}{\raisebox{-0.15em}{\includegraphics[height=1em]{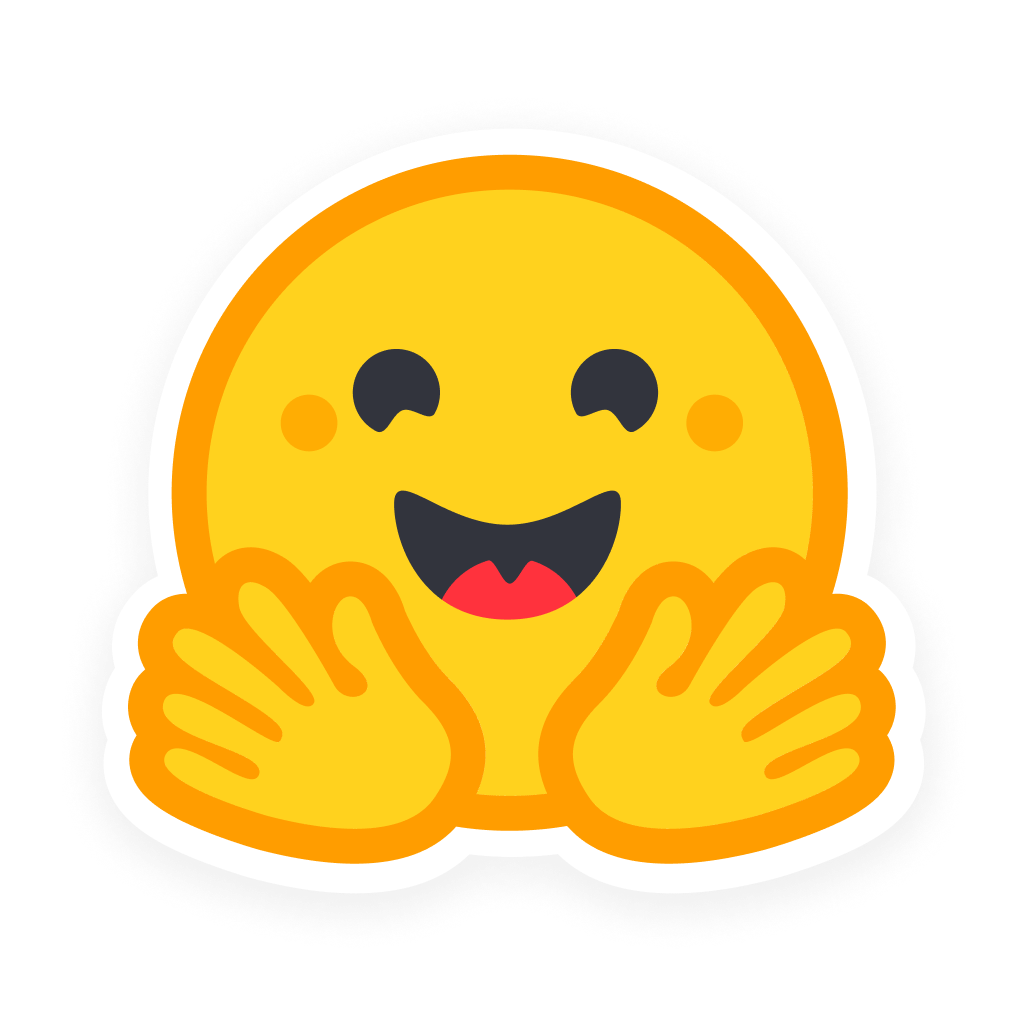}}}

   \usepackage{graphicx}
   \newcommand{\mlogo}[1]{%
     \makebox[1.05em][l]{\raisebox{-0.18em}{%
       \IfFileExists{logos/#1.pdf}{\includegraphics[height=0.95em,width=1.05em,keepaspectratio]{logos/#1.pdf}}{%
       \IfFileExists{logos/#1.png}{\includegraphics[height=0.95em,width=1.05em,keepaspectratio]{logos/#1.png}}{}}}}%
     \hspace{0.3em}}

\title{Seeking in Space: Benchmarking Purposeful Active Exploration of Vision-Language Models for Spatial Reasoning}

\author{
Xiwei Liu$^{1}$,
Yulong Li$^{1}$,
Xuhui Li$^{1}$,
Xinlin Zhuang$^{2}$,
Zhixiang Lu$^{3}$,
Haolin Yang$^{1}$ \\
Jianxu Chen$^{1}$,
Yifan Lu$^{1}$,
Zhiran Zhang$^{1}$,
Yutong Xie$^{1,}$\thanks{Corresponding author.}\\[4pt]
$^{1}$Mohamed bin Zayed University of Artificial Intelligence \\\quad
$^{2}$The Chinese University of Hong Kong \quad
$^{3}$University of Liverpool \\[6pt]
\href{https://xiwei-web.github.io/actspace-bench/}{\faGlobe\ Website} \qquad
\href{https://github.com/Xiwei-web/ActSpace}{\faGithub\ Code} \qquad
\href{https://huggingface.co/datasets/Belmont111/ActSpace-Bench}{\hficon\ Benchmark}
}

\begin{document}
\twocolumn[{%
\renewcommand\twocolumn[1][]{#1}%
\maketitle
\begin{center}
    \centering
    \vspace{-2.0em}                 % 调整图片与作者信息之间的距离
    \includegraphics[width=\textwidth]{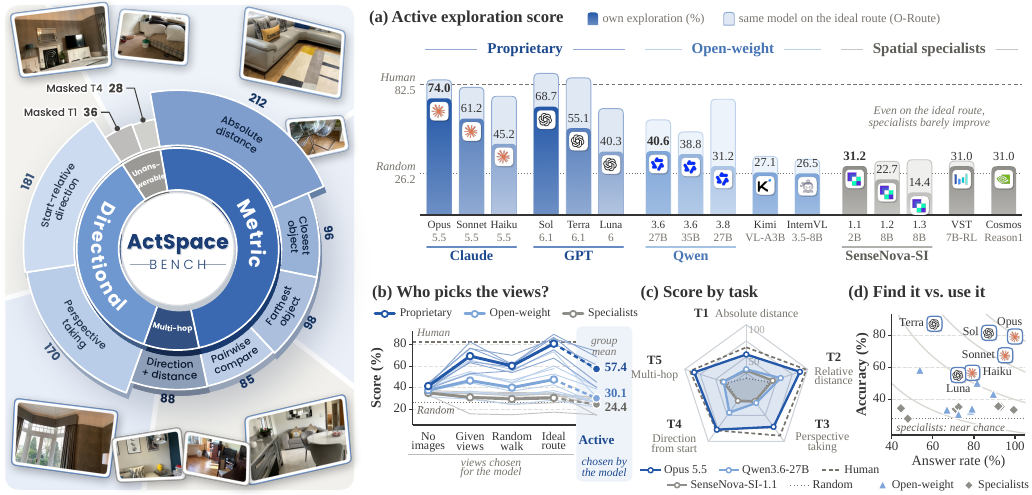}
    \vspace{-2em}                 % 调整图片与图注之间的距离
    \captionof{figure}{\textbf{\textsc{ActSpace-Bench} at a glance.} \textit{Left:} 994 questions in real indoor scenes, where a model must move through the scene to find the evidence. \textit{Right:} (a) Score when each model explores on its own, and when the same model is shown the ideal route. (b) Group-mean score as the views come from different sources. (c) Score by task. (d) How often each model answers and how often answers are correct. }
    \label{fig:teaser}
\end{center}%
% \vspace{0.5em}                      % 调整图注与摘要之间的距离
}]
% \maketitle inside \twocolumn[...] drops the \thanks text; print it here
{\renewcommand{\thefootnote}{\fnsymbol{footnote}}\footnotetext[1]{Corresponding author.}}%

\begin{abstract}
% \vspace{-1em}
Humans explore unfamiliar spaces with purpose, deciding where to look until they have gathered the evidence a spatial question requires. However, can vision–language models (VLMs), which reason increasingly well over curated images and videos, also seek in space when the evidence is not handed to them? We present ActSpace-Bench, a benchmark of 994 questions across 234 exploration episodes in 72 real-scanned indoor scenes, where a model starts from a single view, navigates real captured frames, and must gather evidence before answering. To assess whether such exploration is purposeful, we introduce a controlled evaluation protocol that fixes each model as the reasoner and compares its own exploration against a same-budget random walk and an oracle route, separating evidence seeking from spatial reasoning. Every question is verified to require exploration and answered from 3D annotations, masked-evidence questions test whether models know when to abstain, and over 95\% of sampled questions are confirmed correct by human verification. Evaluations of 23 proprietary, open-weight, and spatially specialized VLMs reveal that the strongest models do seek evidence with purpose but lose much of it to limited memory, whereas weaker and specialized models explore no better than a random walk or fail to reason over the evidence they are given, pointing to memory and evidence-grounded reasoning as the next targets for active spatial intelligence.
\end{abstract}

\vspace{-1em}
\section{Introduction}
\label{sec:intro}
Human spatial understanding is acquired through exploration rather than from a fixed set of views~\cite{yang2025thinking,yang2026mmsibenchbenchmarkmultiimagespatial}. Faced with a spatial question in an unfamiliar environment, people move, turn, and gather the views that the question requires, then stop once the evidence is sufficient. Perception in this sense is an active process, in which what to observe is chosen in service of the task~\cite{wu2024v,liu2025aligning,wang2025actiview}. Vision-language models (VLMs) are increasingly deployed where the same demand arises, as the reasoning core of embodied agents and wearable assistants~\cite{majumdar2024openeqa,kim2024openvla,black2024pi_0,team2025gemini,yang2025egolife}. In such settings no one selects the informative views in advance, and a model must decide for itself where to look and when it has seen enough before it can reason about the space~\cite{ren2024explore,yang20253d}.

Spatial reasoning has become a central target in evaluating VLMs~\cite{cai2025holistic}, and benchmarks have grown from single images~\cite{chen2024spatialvlm,cheng2024spatialrgpt} to multiple views~\cite{yang2026mmsibenchbenchmarkmultiimagespatial,wang2026mindcubespatialmentalmodeling,yeh2026seeing,li2025viewspatialbenchevaluatingmultiperspectivespatial} and egocentric videos~\cite{yang2025thinking}. They have exposed a large gap to human performance, yet share one design choice. The observations are fixed before the model is queried, selected by annotators or recorded by a human who has already walked through the scene. Such benchmarks measure how well a model reasons over evidence it is handed, and leave out how that evidence is obtained. A failure cannot be attributed to missing the evidence or misusing it, and a success says little about whether the model could have found the evidence on its own.

We therefore introduce \textsc{ActSpace-Bench}, a benchmark that returns this choice to the model. In each episode a model is placed at one view of a real indoor scan~\cite{baruch2021arkitscenes,yeshwanth2023scannet++}, moves among the captured frames within a limited number of steps while retaining only its most recent views, and answers when it judges the evidence sufficient or reports that the question cannot be determined. Existing active benchmarks~\cite{majumdar2024openeqa,ren2024explore,jiang2025beyond,sakamoto2026e3vs,hong2026esibenchembodiedspatialintelligence} mostly rely on rendered or simulated scenes and report the active score alone or against a random policy only (Table~\ref{tab:comparison}), which leaves finding and reasoning entangled. We instead pose every question under two matched references, a random walk of the same length and an oracle route through the required evidence. The gap to the random walk shows whether the model explores with purpose, and the gap to the oracle route shows how much score better exploration would add.

\providecolor{pieblue}{HTML}{3F7FC1}
\providecolor{piered}{HTML}{D9604E}
\providecolor{pieyellow}{HTML}{D99A1E}
\providecolor{piegrey}{HTML}{8A8F98}
\providecolor{oursrow}{HTML}{EEF4FB}

\providecommand{\cmark}{\textcolor{pieblue}{\ding{51}}}
\providecommand{\xmark}{\textcolor{piered}{\ding{55}}}
\providecommand{\pmark}{\textcolor{pieyellow}{\ding{51}}}
\providecommand{\na}{\textcolor{piegrey}{--}}
\providecommand{\real}[1]{\textcolor{pieblue}{#1}}   % real captured imagery
\providecommand{\rend}[1]{\textcolor{piered}{#1}}    % rendered / simulated
\providecommand{\sub}[1]{{\scriptsize\textcolor{piegrey}{#1}}}

\providecommand{\vidicon}{%
  \tikz[baseline={([yshift=-0.6ex]current bounding box.center)}, scale=1.35]{%
    \draw[piegrey,line width=0.45pt,rounded corners=0.4pt] (0,0) rectangle (1.6ex,1.2ex);%
    \fill[piegrey] (0.6ex,0.3ex) -- (0.6ex,0.9ex) -- (1.1ex,0.6ex) -- cycle;}}
\providecommand{\imgicon}{%
  \tikz[baseline={([yshift=-0.6ex]current bounding box.center)}, scale=1.35]{%
    \draw[piegrey,line width=0.45pt,rounded corners=0.4pt] (0.3ex,0.3ex) rectangle (1.9ex,1.5ex);%
    \fill[white] (0,0) rectangle (1.6ex,1.2ex);%
    \draw[piegrey,line width=0.45pt,rounded corners=0.4pt] (0,0) rectangle (1.6ex,1.2ex);%
    \fill[piegrey] (0.4ex,0.85ex) circle (0.15ex);%
    \fill[piegrey] (0.15ex,0.12ex) -- (0.7ex,0.7ex) -- (1.05ex,0.4ex) -- (1.45ex,0.85ex) -- (1.45ex,0.12ex) -- cycle;}}
\providecommand{\acticon}{%
  \tikz[baseline={([yshift=-0.6ex]current bounding box.center)}, scale=1.35]{%
    \fill[piegrey] (0.15ex,0.2ex) circle (0.17ex);%
    \draw[piegrey,line width=0.5pt,-{Stealth[length=0.45ex,width=0.5ex]}]
      (0.3ex,0.3ex) .. controls (0.9ex,0.1ex) and (0.6ex,1.1ex) .. (1.7ex,1.0ex);}}

\begin{table}[t]
\centering
\setlength{\tabcolsep}{4.5pt}
\renewcommand{\arraystretch}{1.12}
\resizebox{\columnwidth}{!}{%
\begin{tabular}{l l c c c c}
\toprule
 % & & \textbf{Start-view} & \textbf{Abs-} & \multicolumn{2}{c}{\textbf{Reference}} \\
\cmidrule(lr){5-6}
\textbf{Benchmark} & \textbf{Environment} & \textbf{Start-view} & \textbf{Abstention} & \textbf{Random} & \textbf{Oracle} \\
\midrule
% \multicolumn{6}{l}{\sub{\textit{Passive: observations are given}}} \\
VSI-Bench~\cite{yang2025thinking}          & \real{Real video}  & \na & \xmark & \na & \na \\
MMSI-Bench~\cite{yang2026mmsibenchbenchmarkmultiimagespatial}  & \real{Real images} & \na & \xmark & \na & \na \\
MindCube~\cite{wang2026mindcubespatialmentalmodeling}          & \real{Real images} & \na & \xmark & \na & \na \\
\midrule
% \multicolumn{6}{l}{\sub{\textit{Active: the model collects observations}}} \\
OpenEQA~\cite{majumdar2024openeqa}                             & \rend{Mesh render} & \xmark & \xmark & \xmark & \pmark \\ % Oracle: verify
EXPRESS~\cite{jiang2025beyond}                                 & \rend{Mesh render} & \xmark & \xmark & \xmark & \xmark \\ % Oracle: verify
E3VS-Bench~\cite{sakamoto2026e3vs}                             & \rend{3DGS render} & \pmark & \xmark & \cmark & \xmark \\ % Oracle: verify
ESI-Bench~\cite{hong2026esibenchembodiedspatialintelligence}   & \rend{Simulator}   & \xmark & \xmark & \pmark & \xmark \\ % Oracle: verify
\rowcolor{oursrow}
\textbf{Ours}                                                  & \real{\textbf{Real frames}} & \cmark & \cmark & \cmark & \cmark \\
\bottomrule
\end{tabular}%
}
\vspace{-0.3em}
\caption{\textbf{Comparison with related benchmarks.}
\real{Blue} are real captures and \rend{red} are rendered or simulated.
\pmark: partial support.}
\label{tab:comparison}
\vspace{-1em}
\end{table}

Evaluating 23 proprietary and open VLMs, we find that strong models seek evidence with purpose but are limited by memory, whereas weaker and spatially specialized models fall short in exploration or reasoning over the evidence they are given. When the random walk is restricted to the same recent views, the strongest models outperform it by 12-26 points. Most of this lead disappears once the walk may keep everything it has seen, because evidence found earlier but no longer in view is largely lost. Weaker models do no better than the random walk in either case, and spatially specialized and small open models answer incorrectly even along the oracle route, where the evidence is directly given to them. Our core contributions are \textbf{three-fold}:

\noindent\ding{182} We formulate active spatial reasoning as a problem of seeking evidence, and propose an evaluation protocol that measures whether active exploration is purposeful. By pairing each question with a same-budget random walk and an oracle route, the protocol turns a single active score into separate estimates of exploration and reasoning ability.

\noindent\ding{183} We build \textsc{ActSpace-Bench}, comprising 994 questions over 72 real indoor scenes and five spatial tasks. Every question is verified to require exploration beyond the starting view, and every answer is derived from 3D annotations.

\noindent\ding{184} We provide a diagnosis that locates, for each model, where score is lost among finding, remembering, and reasoning. It identifies memory of gathered evidence and reasoning over it as the main targets for future models.

 \begin{figure*}[t]
    \centering
    \includegraphics[width=\linewidth]{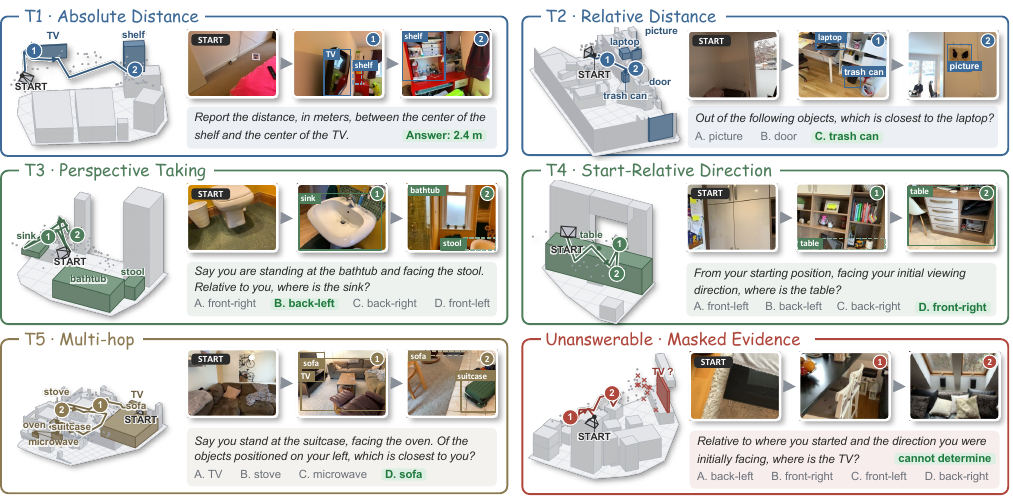}
    \vspace{-2em}
    \caption{Representative ActSpace-Bench samples from each category. Each panel shows the scan with the route and the views along it.}
    \label{fig:cases}
    \vspace{-1em}
\end{figure*}

\begin{figure}[!t]
    \centering
    \includegraphics[width=\linewidth]{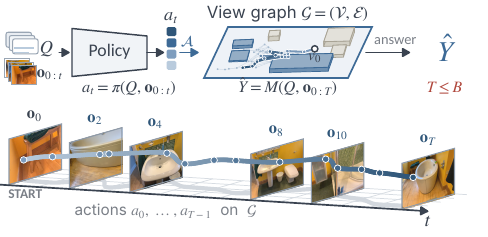}
    \vspace{-2em}
    \caption{Problem formulation.}
    \label{fig:formulation}
    \vspace{-1em}
\end{figure}

%-------------------------------------------------------------------------

\section{Related Work}
\label{sec:related}
\paragraph{Spatial Intelligence in VLMs.}
Benchmarks for spatial intelligence have expanded along the form of their input. Early ones are single-image and emphasize local metric cues such as depth and distance, \eg, SpatialVLM~\cite{chen2024spatialvlm} and SpatialRGPT~\cite{cheng2024spatialrgpt}. Multi-view benchmarks require fusing several images into one layout, including MMSI-Bench~\cite{yang2026mmsibenchbenchmarkmultiimagespatial}, MindCube~\cite{wang2026mindcubespatialmentalmodeling}, All-Angles Bench~\cite{yeh2026seeing}, and ViewSpatial-Bench~\cite{li2025viewspatialbenchevaluatingmultiperspectivespatial}. Video benchmarks extend this to egocentric recordings of whole scenes, as in VSI-Bench~\cite{yang2025thinking}, MMSI-Video-Bench~\cite{lin2025mmsi}, and DSI-Bench~\cite{zhang2025dsi}, and OST-Bench~\cite{lin2026ost} further feeds the frames online to stress memory. In parallel, models have been strengthened with geometry encoders~\cite{fan2026vlm,wu2026spatial} or large-scale spatial tuning~\cite{cai2026scaling,yang2026visual,yang2026cambrian}. In both lines the observations are chosen by an annotator or fixed by a recorded trajectory. \textsc{ActSpace-Bench} keeps the task types and 3D-grounded answers of these benchmarks but lets the model decide what to observe.
\vspace{-2em}
\paragraph{Active Perception and Embodied QA.}
A second line lets the model act before it answers. Within a single image, V$^*$~\cite{wu2024v} and ActiView~\cite{wang2025actiview} study visual search through zooming and shifting the field of view. In 3D scenes, embodied QA benchmarks such as OpenEQA~\cite{majumdar2024openeqa}, EXPRESS-Bench~\cite{jiang2025beyond}, E3VS-Bench~\cite{sakamoto2026e3vs}, and ESI-Bench~\cite{hong2026esibenchembodiedspatialintelligence} require navigation in rendered or simulated environments, while methods such as Explore until Confident~\cite{ren2024explore} and 3D-Mem~\cite{yang20253d} improve exploration with semantic maps and scene memory. These works establish that active exploration is difficult, yet they seldom pair the active score with both a chance-level and an oracle reference, leaving exploration and reasoning entangled. \textsc{ActSpace-Bench} provides both references on real captures and tests whether a model recognizes absent evidence.

% \clearpage
% \newpage
\section{ActSpace-Bench}
\label{sec:method}
\begin{figure*}[!t]
    \centering
    \includegraphics[width=\linewidth]{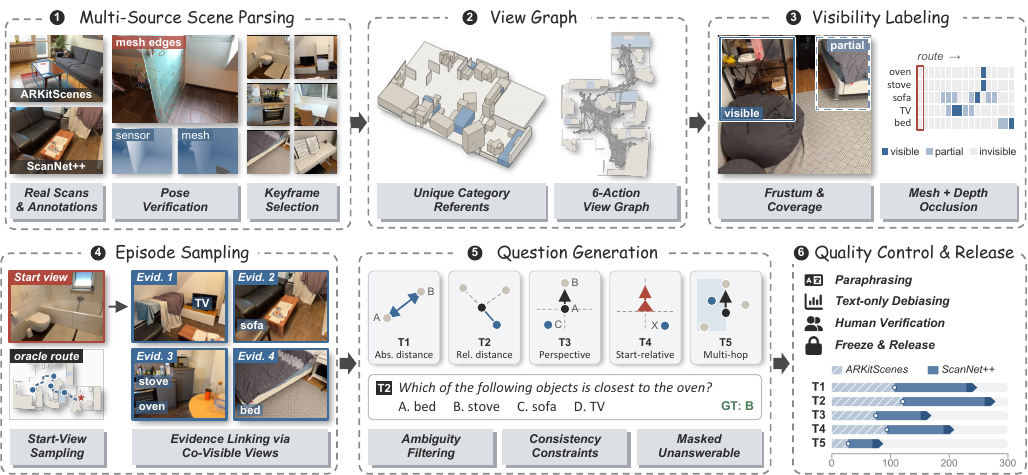}
    \vspace{-2em}
    \caption{Overview of the ActSpace-Bench construction pipeline.}
    \label{fig:pipeline}
    \vspace{-1em}
\end{figure*}

% ------------------------------------------------------------

\subsection{Problem Formulation}
\label{sec:formulation}

We model active spatial reasoning as answering questions about a scene while choosing what to observe. As shown in Figure~\ref{fig:formulation}, a scene is a view graph $\mathcal{G}=(\mathcal{V},\mathcal{E})$ whose nodes are real captured frames and whose edges correspond to a set of movement actions $\mathcal{A}$. An episode begins at node $v_0$ with a question episode $Q$. At step $t$, the agent holds observations $o_{0:t}$, where each $o_i$ is the frame at node $v_i$ together with its pose relative to $v_0$, and selects
\begin{equation}
a_t=\pi(Q,o_{0:t}),\quad a_t\in\mathcal{A}\cup\{\texttt{answer}\}.
\setlength\abovedisplayskip{4pt}%
\setlength\belowdisplayskip{4pt}%
\end{equation}
Exploration stops when the agent chooses to answer or after $B$ steps, and the model $M$ then answers all questions from the observations collected along its trajectory,
\begin{equation}
\hat{Y}=M(Q,o_{0:T}),\quad T\le B.
\setlength\abovedisplayskip{4pt}%
\setlength\belowdisplayskip{4pt}%
\end{equation}
Ground-truth answers $Y^\star$ come from 3D annotations. Each question is built so that the start view does not show all objects involved, while some path within $B$ steps does.

This formulation separates collecting evidence ($\pi$) from reasoning over it ($M$). Let $S(\pi)$ be the expected score of $M$ when its observations come from policy $\pi$. In the active setting, $M$ itself acts as the policy $\pi_M$. Keeping $M$ fixed, we replace its policy with a random walk $\pi_{\text{rw}}$ under the same budget, or with the oracle route $\pi_{\text{or}}$ that observes all required objects in the fewest steps, and define
\begin{equation}
\Delta_{\mathrm{RW}}=S(\pi_M)-S(\pi_{\text{rw}}),\quad
\Delta_{\mathrm{O}}=S(\pi_{\text{or}})-S(\pi_M).
\setlength\abovedisplayskip{4pt}%
\setlength\belowdisplayskip{4pt}%
\end{equation}
A positive $\Delta_{\mathrm{RW}}$ indicates purposeful exploration beyond chance coverage, and $\Delta_{\mathrm{O}}$ is the accuracy lost when the model has to collect the required evidence by itself.

% ------------------------------------------------------------
\subsection{Task Taxonomy}
\label{sec:taxonomy}
Answering spatial questions in an unfamiliar scene requires assembling a coherent spatial representation from egocentric views gathered at different places. Our taxonomy probes this representation along two complementary axes: \textit{Metric}, what it encodes, and \textit{Directional}, the reference frame in which it is queried (samples in Figure~\ref{fig:cases}).

On the metric axis, \textbf{absolute distance} (T1) requires recovering metric scale across views, whereas \textbf{relative distance} (T2) only requires ordering distances consistently, which separates failures of scale estimation from failures of relational structure. On the directional axis, the reference frame moves from the scene to the agent. \textbf{Perspective taking} (T3) queries an allocentric frame anchored at objects, which the agent must imagine occupying. \textbf{Start-relative direction} (T4) asks where a target lies relative to the agent's initial pose after it has moved, so a later observation must be transformed back into the starting frame. Passive benchmarks cannot pose this question, as their observer never moves. \textbf{Multi-hop questions} (T5) chain the two axes by filtering objects by direction before comparing distances, and test composed queries rather than isolated lookups.

We further adapt the questions to the active setting in two ways. Instead of asking one question per exploration, we ask several questions about different objects after a single exploration, so the agent has to cover multiple targets. Some of these questions are logically linked, which lets us check whether its answers agree with each other. We also include questions whose evidence has been removed from the scene. Since the agent alone decides when it has seen enough, recognizing that the required evidence does not exist matters as much as reasoning over evidence that does.

% ------------------------------------------------------------
\subsection{Construction Pipeline}
\label{sec:pipeline}
Figure~\ref{fig:pipeline} summarizes our benchmark construction pipeline. \textsc{ActSpace-Bench} pairs a geometry-driven generator, which derives every view, visibility label, and answer from aligned 3D scans, with iterative human verification whose findings were used to revise the construction rules.

\vspace{-1em}
\paragraph{Scene Parsing.}
We build on ARKitScenes~\cite{baruch2021arkitscenes} and ScanNet++~\cite{yeshwanth2023scannet++}, which provide captured RGB-D frames, camera trajectories, meshes, and object annotations for the same scenes. The agent therefore observes raw captured images, while every quantity behind the questions is computed from aligned 3D data. We verify each camera pose by comparing mesh-rendered depth with sensor depth, discard frames that disagree, and select sharp keyframes along the trajectory.
Objects are referred to only by unique category names in the scene, so every question has an unambiguous referent.

\vspace{-1em}
\paragraph{View Graph.}
Keyframes form the nodes of the view graph, and each action links a node to the closest keyframe consistent with that motion relative to the current heading. We keep the largest strongly connected component, so every view the agent reaches is also a view it can leave.

\vspace{-1em}
\paragraph{Visibility Labeling.}
For every object and keyframe, we project the object's surface points into the image and test occlusion against both mesh and sensor depth, correcting for mesh holes and scene changes such as opened doors. Each pair is labeled visible, partial, or invisible, where partial neither confirms nor rules out recognition. This visibility map underlies all later stages and our analysis of exploration.
\vspace{-1em}
\paragraph{Episode Sampling.}
Each episode is built around a cluster of objects that cannot all be seen in a single view. We sample a start view from which most of them are invisible, and require every evidence object to be clearly visible from some node. Objects within a question must also be connected through views in which they co-occur, so the queried relations can actually be inferred from images. These views define the oracle route used as our upper reference.
\newcommand{\tr}{\rlap{$^{\ddagger}$}}

\begin{table*}[t]
\centering
\caption{\textbf{Main results on ActSpace-Bench} (\%). The human reference answers from oracle views rather than by exploring. \best{Bold} and \second{shaded} mark the best and second-best results. $^{\ddagger}$: over 15\% of the responses in this setting are truncated or cut off by context overflow (Appendix~\ref{app:hyper}); such settings are excluded from the ranking, and gaps computed from them are omitted. P-40: Passive-40.}
\vspace{-0.8em}
\label{tab:main}
\vspace{2pt}
\footnotesize
\setlength{\tabcolsep}{3.1pt}
\renewcommand{\arraystretch}{1.05}
\setlength{\aboverulesep}{0pt}
\setlength{\belowrulesep}{0pt}
% --- group strip: grey vertical tab on the left carrying the rotated group name
\providecolor{asGroup}{gray}{0.90}          % strip fill (reuse your \grouprow grey if you like)
\providecolor{asGroupRule}{gray}{0.62}      % thin rule between groups
% put \grouplabel{n}{text} in the LAST row of a group (n = number of rows in it)
\providecommand{\grouplabel}[2]{\multirow{-#1}{*}{\rotatebox[origin=c]{90}{\scriptsize\bfseries\itshape #2}}}
\providecommand{\groupsep}{\arrayrulecolor{asGroupRule}\specialrule{0.3pt}{0pt}{0pt}\arrayrulecolor{black}}
\renewcommand{\grouplabel}[2]{\multirow{-#1}{*}{\rotatebox[origin=c]{90}{\scriptsize\itshape\color{black!70} #2}}}
% tabular* stretches the table to exactly \textwidth; the stretch is switched off
% inside the shaded Active block so the shading stays continuous.
\begin{tabular*}{\textwidth}{@{}>{\columncolor{asGroup}[2pt][2pt]}c@{\hspace{5pt}}l @{\extracolsep{\fill}} cccc
  !{\extracolsep{0pt}}>{\columncolor{asActive}}c>{\columncolor{asActive}}c>{\columncolor{asActive}}c>{\columncolor{asActive}}c
  !{\extracolsep{\fill}} cc ccccc@{}}
\toprule
\multicolumn{1}{@{}c@{\hspace{5pt}}}{} & \multirow{2}{*}{\textbf{Model}}
 & \multicolumn{4}{c}{\textbf{Reference settings}}
 & \multicolumn{4}{>{\columncolor{asActive}}c}{\textbf{Active (main setting)}}
 & \multicolumn{2}{c}{\textbf{Exploration}}
 & \multicolumn{5}{c}{\textbf{Active score by task}} \\
\cmidrule(lr){3-6}\cmidrule(lr){7-10}\cmidrule(lr){11-12}\cmidrule(lr){13-17}
\multicolumn{1}{@{}c@{\hspace{5pt}}}{} & & Blind & P-40 & RW-40 & O-Route
 & Score & AR & Sel. & Rej.
 & $\Delta_{\mathrm{RW}}$ & $\Delta_{\mathrm{O}}$
 & T1 & T2 & T3 & T4 & T5 \\
\midrule
% ------------------------------------------------------------------ Base
 & \mlogo{}Random Guessing              & 28.0 & 26.2 & 26.2 & 26.2 & 26.2 & \na & 28.0 & \na & \na & \na & 17.0 & 37.1 & 25.0 & 21.7 & 31.3 \\
\grouplabel{2}{Base} & \mlogo{}Human (oracle views)  & \na & \na & \na & \na & 82.5 & 98.4 & 83.9 & 81.8 & \na & \na & 65.0 & 98.1 & 88.9 & 78.9 & 91.7 \\
\groupsep
% ------------------------------------------------------------------ Proprietary
 & \mlogo{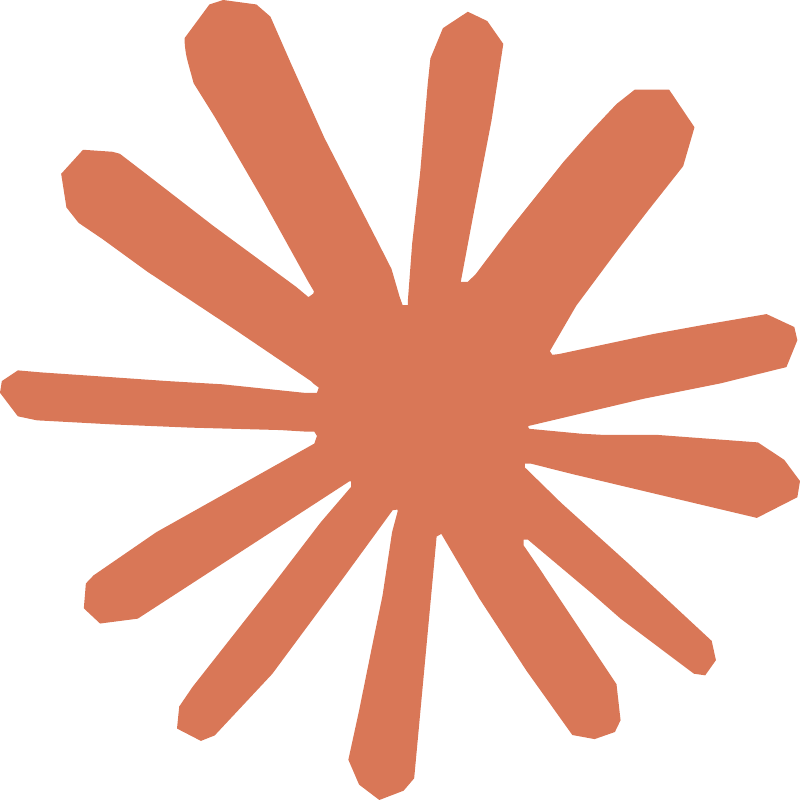}Claude Opus 5.5       & \second{44.1} & 76.4 & \best{70.1} & 85.6 & \best{74.0} & 100 & 79.2 & 0.0 & \textcolor{NiceGreen}{+3.9} & 11.6 & \best{54.0} & \best{87.5} & \best{72.0} & \best{77.5} & \best{84.8} \\
 & \mlogo{claude}Claude Sonnet 5.5     & 38.6 & 64.2 & 59.2 & 80.9 & 61.2 & 95.3 & 67.5 & 15.9 & \textcolor{NiceGreen}{+2.0} & 19.7 & 48.1 & 75.3 & 53.7 & 66.2 & 55.7 \\
 & \mlogo{claude}Claude Haiku 5.5      & 39.0 & 53.0 & 55.6 & 75.3 & 45.2 & 79.2 & 56.7 & \second{50.3} & \gain{-10.4} & 30.1 & 37.7 & 55.6 & 30.1 & 57.5 & 32.5 \\
% & \mlogo{openai}GPT-6 Astra           & 40.3 & 83.7 & 64.6 & 91.3 & 71.7 & 89.6 & 79.5 & 18.9 &  & \na & \na & \na & \na & \na & \na \\
 & \mlogo{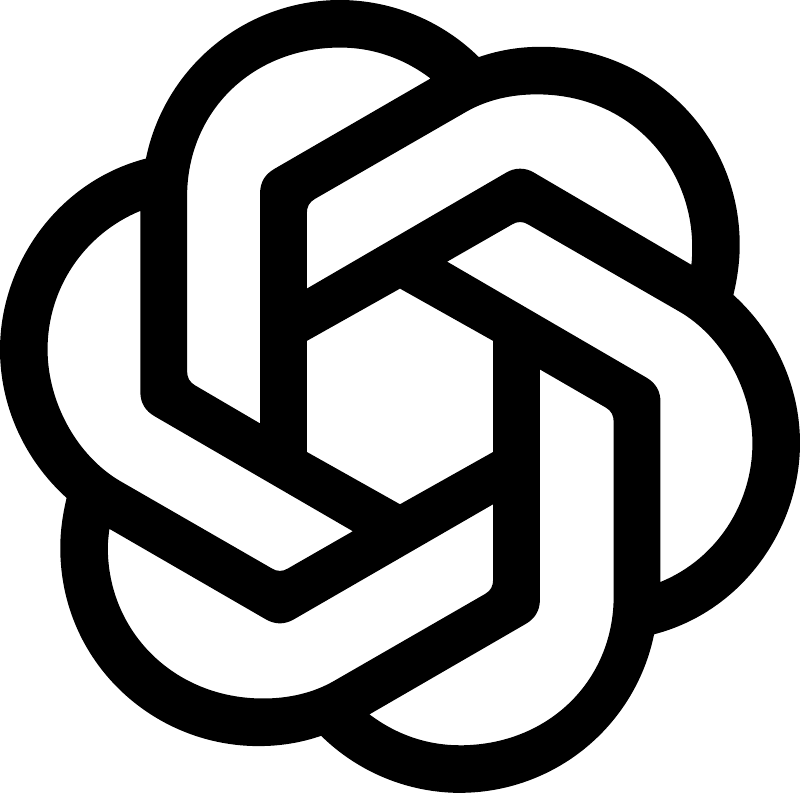}GPT-6.1 Sol           & 42.1 & \best{82.6} & \second{61.8} & \best{89.7} & \second{68.7} & 87.4 & \second{81.5} & 33.3 & \textcolor{NiceGreen}{+6.9} & 21.0 & \second{52.6} & \second{76.8} & \second{71.3} & \second{77.0} & \second{64.6} \\
 & \mlogo{openai}GPT-6.1 Terra         & \best{44.2} & \second{76.9} & 61.4 & \second{86.9} & 55.1 & 60.8 & \best{87.1} & \best{84.1} & \gain{-6.3} & 31.8 & 47.4 & 59.0 & 44.5 & 69.1 & 50.6 \\
\grouplabel{6}{Proprietary} & \mlogo{openai}GPT-6 Luna            & 40.9 & 63.3 & 54.0 & 67.5 & 40.3 & 72.4 & 55.3 & 44.4 & \gain{-13.7} & 27.2 & 35.6 & 57.9 & 20.7 & 43.6 & 26.6 \\
\groupsep
% ------------------------------------------------------------------ Open-weight general
 & \mlogo{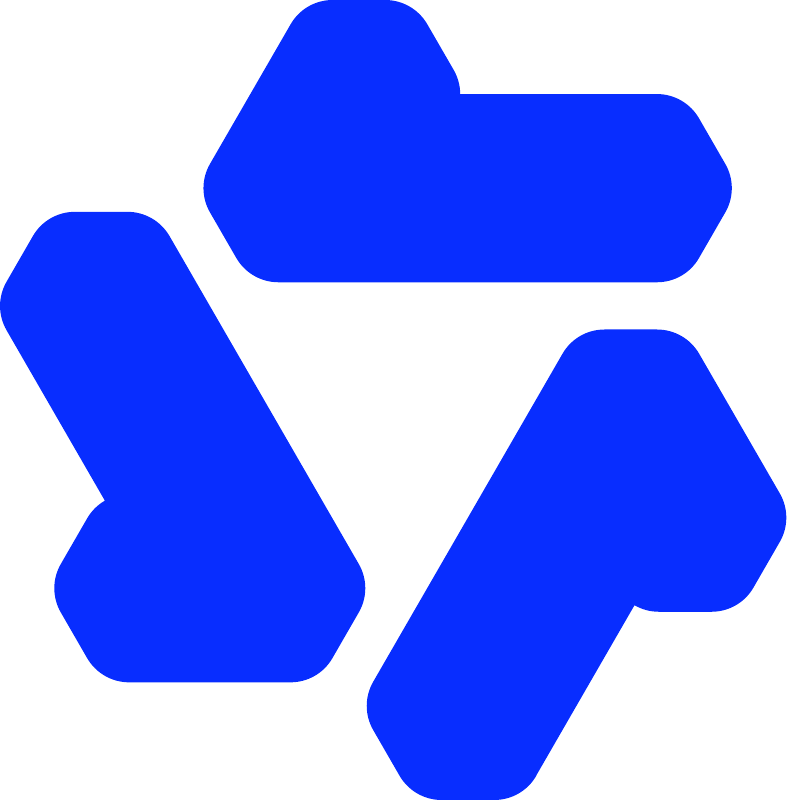}Qwen3.5-4B             & 36.0 & 29.7\tr & 27.3\tr & 40.3 & 17.2\tr & 50.5 & 32.6 & 27.0 & \na & \na & 10.3 & 21.4 & 13.4 & 20.6 & 22.8 \\
 & \mlogo{qwen}Qwen3.5-9B             & 37.9 & 31.0\tr & 15.4\tr & 33.9\tr & 23.2\tr & 50.3 & 40.6 & 63.5 & \na & \na & 17.0 & 26.6 & 17.1 & 30.4 & 25.3 \\
 & \mlogo{qwen}Qwen3.6-27B            & \best{42.9} & \second{55.0} & 44.9 & \second{60.3} & \best{40.6} & 81.6 & \second{50.4} & 31.7 & \gain{-4.3} & 19.7 & \second{30.9} & \best{56.1} & \second{26.2} & \best{44.6} & \best{36.7} \\
 & \mlogo{qwen}Qwen3.6-35B-A3B        & 40.2 & 53.5 & 44.5 & 52.7 & \second{38.8} & 89.5 & 43.5 & \second{36.5} & \gain{-5.7} & 13.9 & \best{32.9} & \second{54.2} & \best{26.8} & 37.7 & \second{31.6} \\
 & \mlogo{qwen}Qwen3.8-27B            & \second{42.1} & \best{65.7} & \best{56.9} & \best{73.4} & 31.2 & 53.7 & \best{58.4} & 28.6 & \gain{-25.7} & 42.2 & 25.5 & 42.1 & 16.5 & \second{38.2} & 24.1 \\
 & \mlogo{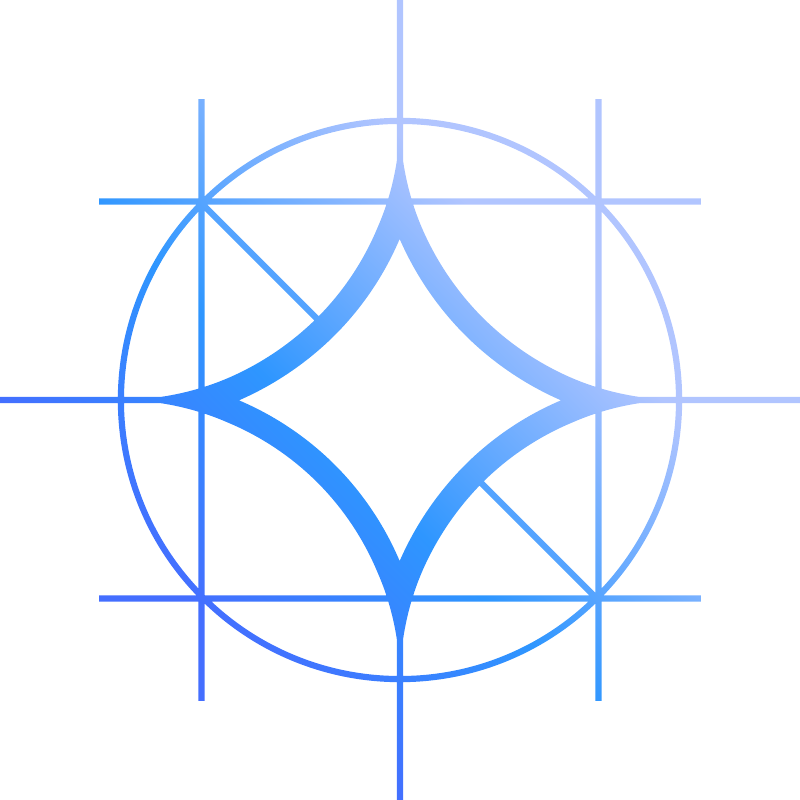}Gemma-4-31B-it        & 40.2 & 52.5 & \second{47.0} & 57.7 & 33.4\tr & 62.8 & 53.8 & 27.0 & \na & \na & 18.9 & 47.6 & 26.2 & 39.7 & 27.8 \\
 & \mlogo{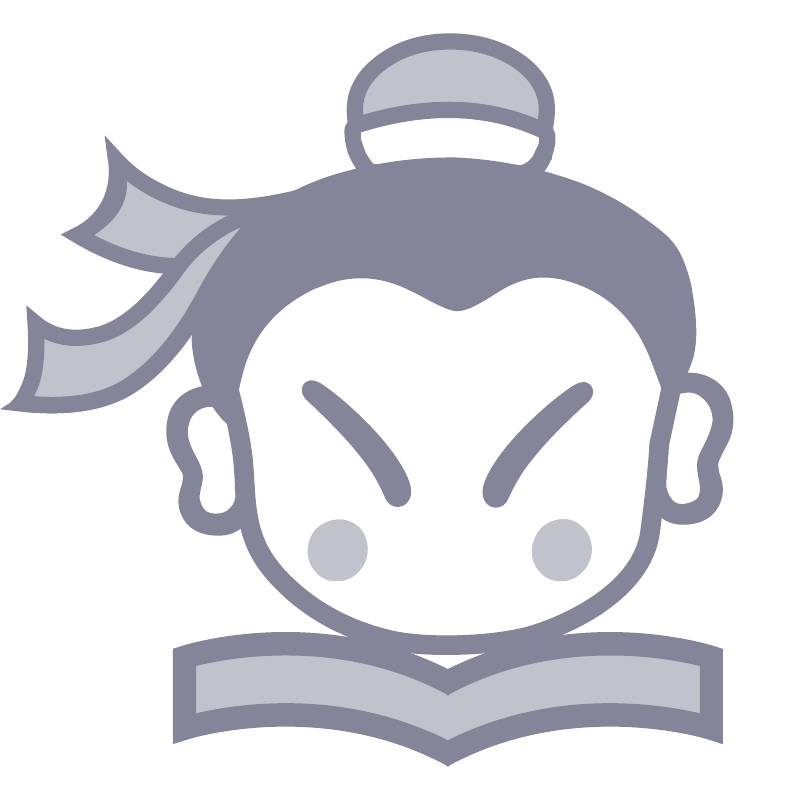}InternVL3.5-8B     & 36.0 & 35.3 & 33.5 & 35.3 & 26.5 & 78.4 & 33.1 & 34.9 & \gain{-7.0} & 8.8 & 24.7 & 32.1 & 20.1 & 25.5 & 29.1 \\
 & \mlogo{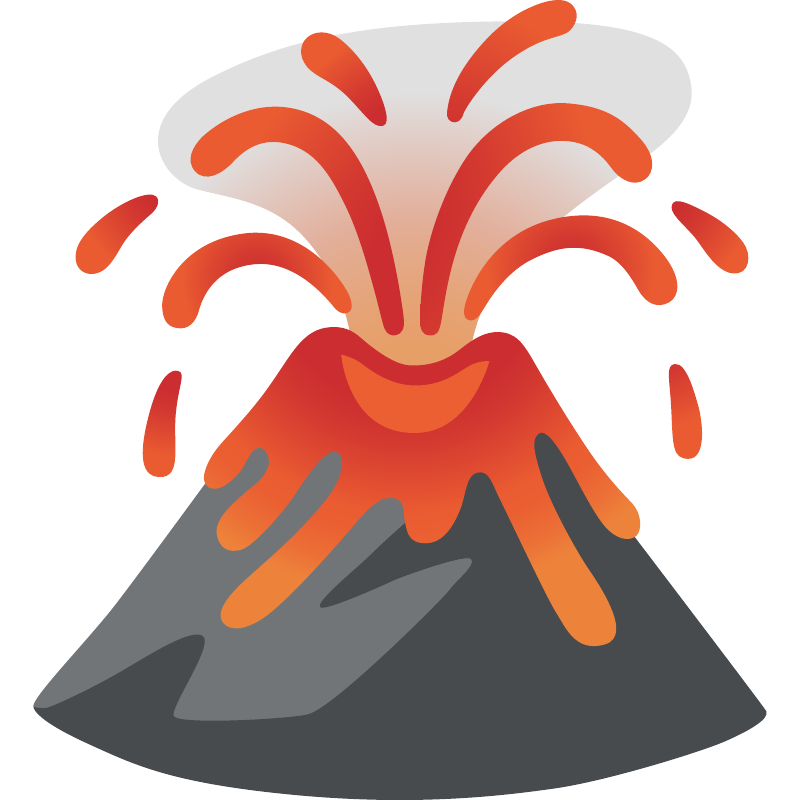}LLaVA-OneVision-7B    & 31.6 & 29.2 & 24.3 & 28.4 & 21.7 & 72.5 & 31.0 & 11.1 & \textcolor{NiceRed}{$-2.6$} & 6.7 & 13.1 & 33.2 & 15.2 & 21.1 & 24.1 \\
 & \mlogo{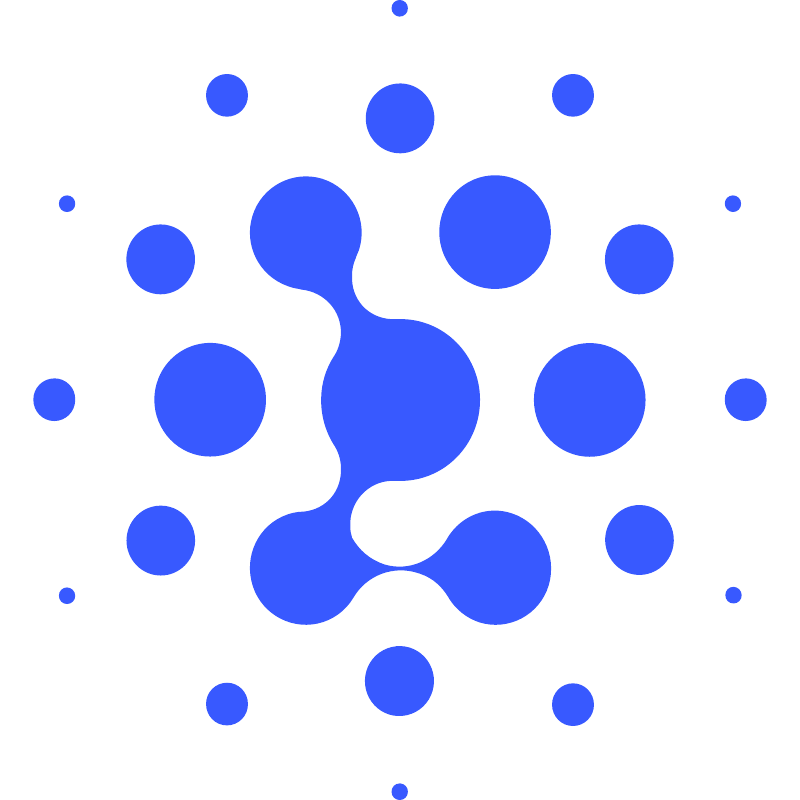}GLM-4.6V-Flash          & 36.9 & 44.7 & 36.7 & 42.7 & 25.0 & 66.9 & 33.7 & \best{60.3} & \gain{-11.7} & 17.7 & 21.6 & 33.9 & 20.1 & 23.0 & 20.3 \\
\grouplabel{10}{Open-weight general} & \mlogo{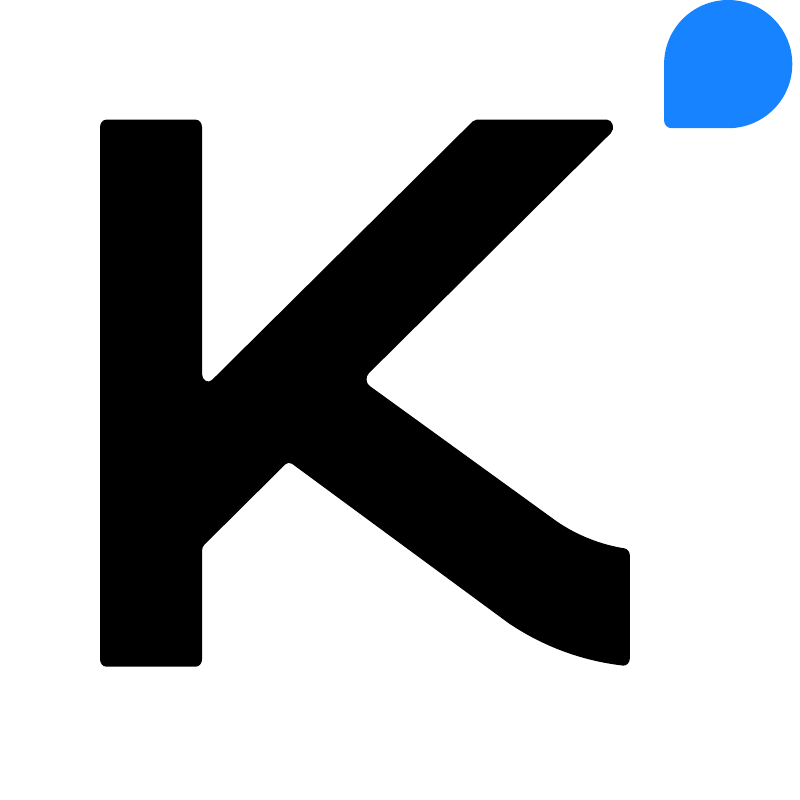}Kimi-VL-A3B-Thinking   & 33.1 & 36.8 & 33.1 & 37.2 & 27.1 & 79.1 & 34.4 & 25.4 & \gain{-6.0} & 10.1 & 19.9 & 36.2 & 25.0 & 24.0 & 30.4 \\
% & \mlogo{minicpm}MiniCPM-V-4.5   & \na & \na & \na & \na & \na & \na & \na & \na & \na & \na & \na & \na & \na & \na & \na \\
\groupsep
% ------------------------------------------------------------------ Spatial specialist
 & \mlogo{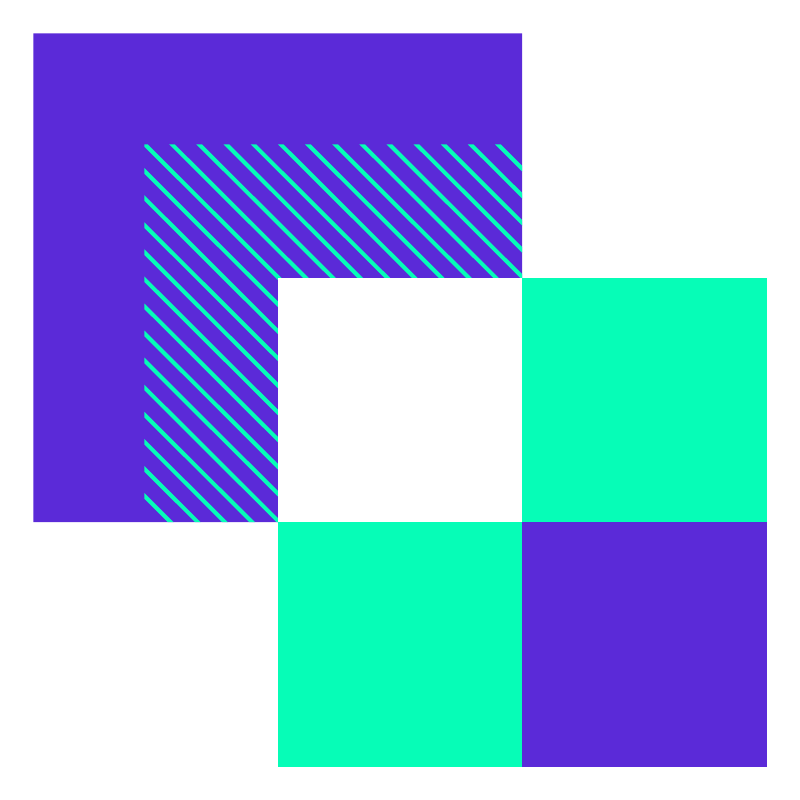}SenseNova-SI-1.1-2B   & 35.9 & 31.7 & 31.1 & 31.9 & \best{31.2} & 99.4 & 33.6 & 0.0 & \textcolor{NiceGreen}{+0.1} & 0.7 & \best{29.3} & 43.2 & 22.6 & 22.1 & \best{38.0} \\
 & \mlogo{sensenova}SenseNova-SI-1.2-8B   & 34.9 & 34.5 & 33.2 & 34.1 & 22.7 & 71.1 & 34.2 & 0.0 & \gain{-10.5} & 11.4 & 23.2 & 31.7 & 20.7 & 13.7 & 17.7 \\
 & \mlogo{sensenova}SenseNova-SI-1.3-8B   & 34.1 & 34.3 & \second{34.6} & \best{35.1} & 14.4 & 44.5 & 34.6 & 0.0 & \gain{-20.2} & 20.7 & 14.1 & 20.3 & 11.6 & 10.8 & 10.1 \\
 & \mlogo{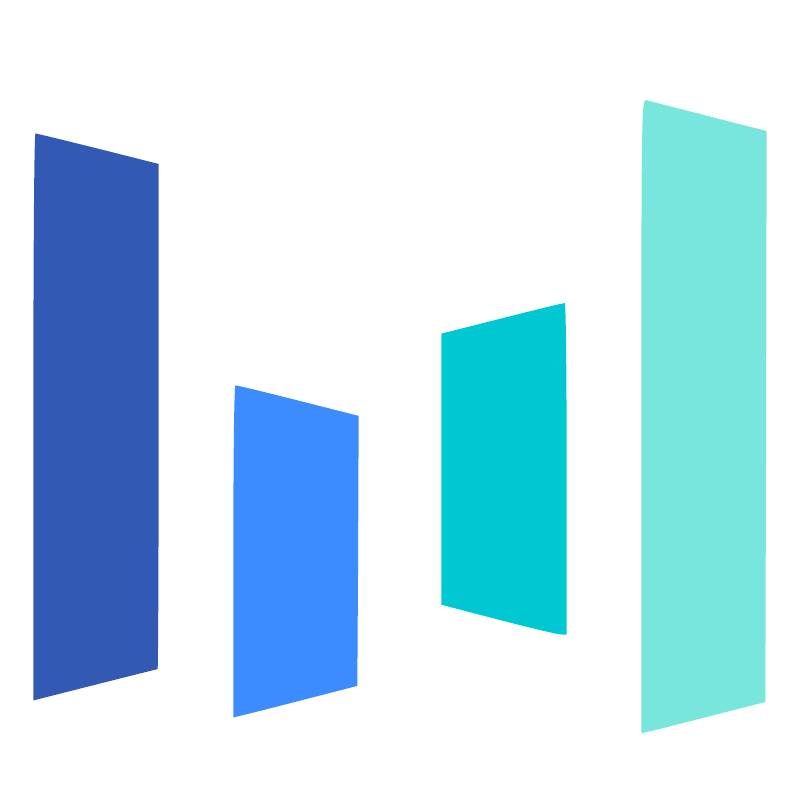}VST-7B-RL                   & \second{36.1} & \second{34.7} & 32.4 & 34.2 & \second{31.0} & 93.1 & \second{35.6} & 1.6 & \gain{-1.4} & 3.2 & \second{28.9} & \best{45.8} & 24.4 & 19.6 & 30.4 \\
 & \mlogo{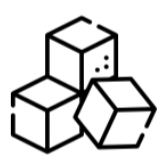}Cambrian-S-7B          & 33.2 & 15.6 & 14.8 & 17.2 & 14.7 & 47.8 & 28.1 & \best{31.7} & \gain{-0.1} & 2.5 & 9.8 & 17.0 & 12.8 & 16.7 & 20.3 \\
 & \mlogo{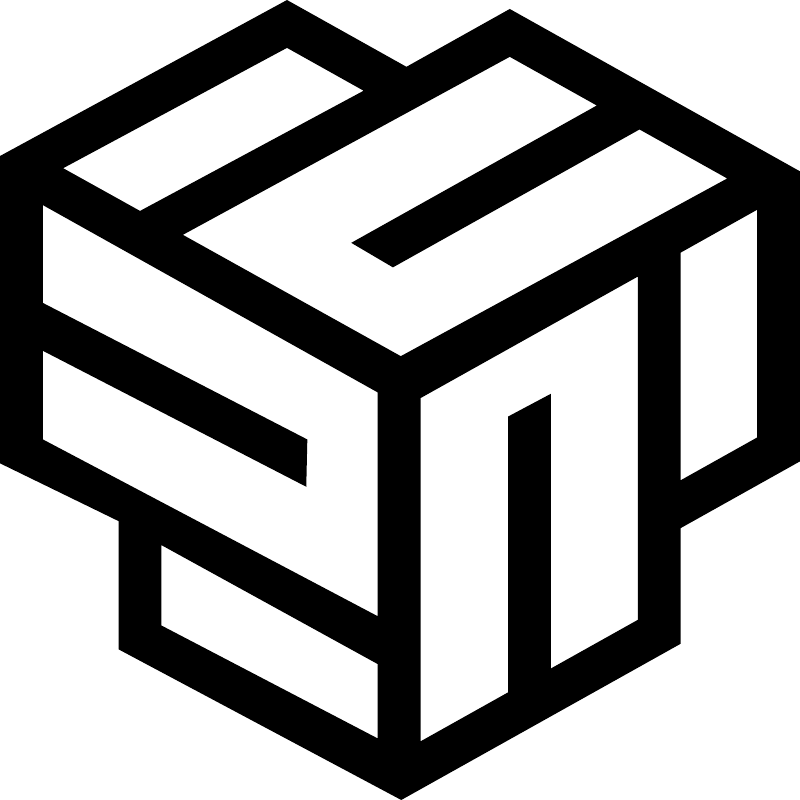}RoboBrain2.0-7B       & \best{37.3} & \best{36.1} & \best{35.5} & \second{34.9} & 26.1 & 72.6 & 35.4 & \best{31.7} & \gain{-9.4} & 8.8 & 19.3 & 32.1 & \best{28.0} & \second{23.5} & 29.1 \\
\grouplabel{7}{Spatial specialist} & \mlogo{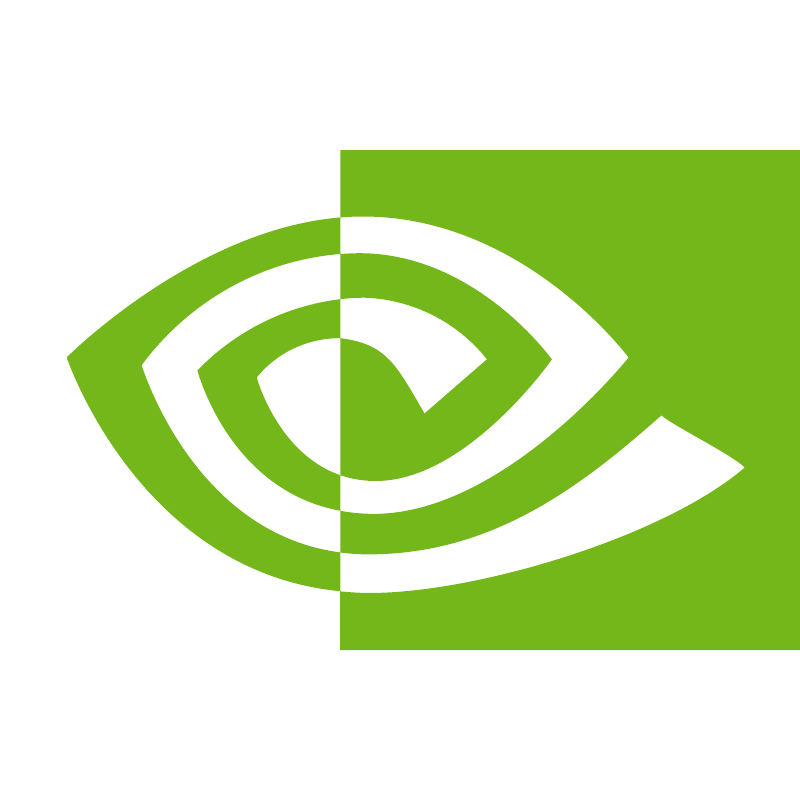}Cosmos-Reason1-7B        & 34.4 & 29.9 & 24.5 & 25.9 & \second{31.0} & 91.9 & \best{35.8} & \second{3.2} & \textcolor{NiceGreen}{+6.5} & $-$5.1 & 22.9 & \second{44.3} & \second{26.2} & \best{26.0} & \second{32.9} \\
\bottomrule
\end{tabular*}
\vspace{-1em}
\end{table*}

% \begin{figure*}[h]
%     \centering
%     \includegraphics[width=\linewidth]{figures/exploration_abstention_new_muted.pdf}
%     \vspace{-2em}
%     \caption{\textbf{Exploration and abstention behavior.}
% (a) Episodes with all evidence seen against median steps used; Gray curve: random walk.
% (b) Stopping step (marker: median, bar: interquartile range).
% (c) Gain in evidence found over a random walk of equal steps; filled markers have 95\% intervals excluding zero.
% (d) Abstention (\texttt{cannot\_determine} on answerable questions).
% $^{\ddagger}$: affected by truncation.}
%     \label{fig:behavior}
%     \vspace{-1em}
% \end{figure*}
\vspace{-1em}
\paragraph{Question Generation.}
Questions are instantiated from templates, and their answers are computed from object centers and the start pose. We discard questions whose answers lie near a decision boundary and add logically linked questions where possible. Unanswerable questions are created by masking every view of a target while keeping the rest of its group answerable and the view graph connected.
\vspace{-1em}
\paragraph{Quality Control.}
Question wording is sampled from paraphrases of each template. A text-only classifier trained and tested on disjoint scenes then guides iterative pruning until no question type is answerable above chance from text alone. Three expert annotators answers sampled episodes from the evidence views without the ground truth and afterwards checks each answer against object boxes and a top-down map. Disagreements are fed back into the pipeline.
% ------------------------------------------------------------
% ---------------------------------------------------------------
% Preamble additions (main.tex):

% Compile twice (arrows use remember picture).
% ---------------------------------------------------------------
\providecolor{stBlue}{HTML}{3F7FC1}
\providecolor{stGreen}{HTML}{2E8B57}
\providecolor{stAmber}{HTML}{C98A12}
\providecolor{stGrey}{HTML}{8A8F98}
\providecolor{stRow}{HTML}{EEF4FB}
\providecommand{\cmark}{\textcolor{stBlue}{\ding{51}}}
\providecommand{\xmark}{\textcolor{stGrey}{\ding{55}}}
% coloured label:  \stag{colour}{text}
\providecommand{\stag}[2]{\textcolor{#1!85!black}{\scriptsize\bfseries #2}}
\providecommand{\smark}[1]{\tikz[remember picture,baseline=-0.5ex]\coordinate(#1);}

\begin{table}[t]
\centering
\caption{\textbf{Evaluation settings.} All settings use the same questions and differ only in how the visual evidence is obtained and who chooses it. $\Delta_{\mathrm{RW}}$ (Active $-$ RW-40) measures the gain of the model's own exploration over a random policy, and $\Delta_{\mathrm{O}}$ (O-Route $-$ Active) measures how far it remains from an oracle route.}
\vspace{-0.8em}
\footnotesize
\setlength{\tabcolsep}{3.3pt}
\renewcommand{\arraystretch}{1.05}
\begin{tabular}{@{}l l c c c@{\hspace{27pt}}}
\toprule
\textbf{Setting} & \textbf{Evidence} & \textbf{Pose} & \textbf{Chosen by} & \textbf{Turns} \\
\midrule
Blind      & none                        & \xmark & \stag{stGrey}{none}    & 1 \\
Passive-40 & 40 random frames            & \xmark & {random} & 1 \\
RW-40      & 40-step random walk         & \cmark & {random} & 1\smark{rw} \\
\rowcolor{stRow}
\textbf{Active} & \textbf{own exploration (${\le}$40)} & \cmark & \stag{stBlue}{model} & multi\smark{ac} \\
O-Route    & shortest evidence route     & \cmark & {oracle} & 1\smark{or} \\
\bottomrule
\end{tabular}
\begin{tikzpicture}[remember picture,overlay,>={Stealth[length=3pt]},
                    lab/.style={font=\scriptsize,inner sep=1pt,anchor=west}]
  \draw[->,stBlue,line width=0.6pt] ([xshift=6pt]rw) to[out=0,in=0,looseness=1.6]
        node[lab,text=stBlue]{$\Delta_{\mathrm{RW}}$} ([xshift=6pt,yshift=1pt]ac);
  \draw[->,stGreen,line width=0.6pt] ([xshift=6pt,yshift=-1pt]ac) to[out=0,in=0,looseness=1.6]
        node[lab,text=stGreen]{$\Delta_{\mathrm{O}}$} ([xshift=6pt]or);
\end{tikzpicture}
\vspace{-1em}
\label{tab:settings}
\vspace{-0.8em}
\end{table}
\begin{figure*}[t]
    \centering
    \includegraphics[width=\linewidth]{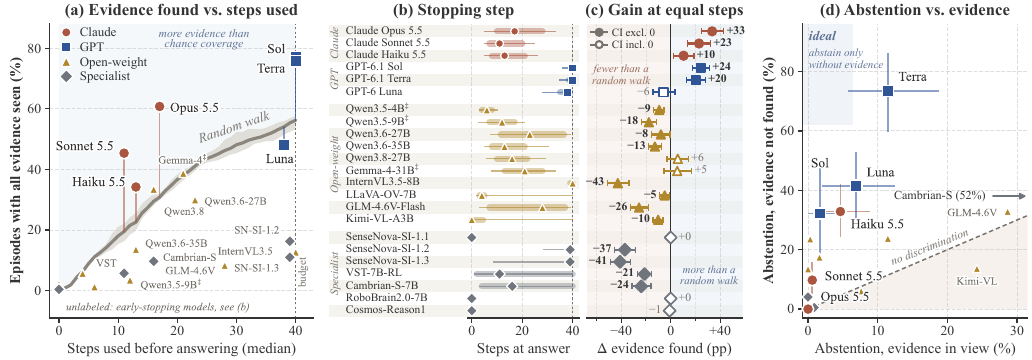}
    \vspace{-2em}
    \caption{\textbf{Exploration and abstention behavior.}
(a) Episodes with all evidence seen against median steps used; Gray curve: random walk.
(b) Stopping step (marker: median, bar: interquartile range).
(c) Gain in evidence found over a random walk of equal steps; filled markers have 95\% intervals excluding zero.
(d) Abstention (\texttt{cannot\_determine} on answerable questions).
$^{\ddagger}$: affected by truncation.}
    \label{fig:behavior}
    \vspace{-1em}
\end{figure*}

\section{Evaluation Protocol}
\label{sec:protocol}
Our protocol is built around a single question, whether a model's own exploration gathers better evidence than a random walk, and every setting below serves to answer it.
\vspace{-1em}
\paragraph{Active Setting.}
The model receives the start image and all questions of the episode, then issues one action per turn. After each move it receives the new frame, its displacement since the previous step, and its accumulated pose relative to the start, mimicking a walker's sense of self-motion. The start image and the eight most recent frames stay in context, while the full history of actions and poses is kept as text. Each episode allows $B{=}40$ steps, failed moves included, enough to cover the shortest evidence route of nearly every episode. The model ends the episode by answering all questions at once and may reply \texttt{cannot\_determine}.

\vspace{-1em}
\paragraph{Reference Settings.}
Table~\ref{tab:settings} orders the reference settings by how much structure their evidence carries. Blind measures text priors, and Passive-40 offers broad but unordered coverage of the scene. RW-40 and O-Route present posed views along a trajectory, as in the active setting, chosen by a random walk or by an oracle following the shortest evidence route. The references see all their views in a single turn, whereas the active agent retains only recent images.

\vspace{-1em}
\paragraph{Metrics.}
Multiple-choice questions are scored by accuracy, and T1 by the mean relative accuracy of VSI-Bench~\cite{yang2025thinking} over thresholds $\theta\in\{0.50,0.55,\dots,0.95\}$,
\begin{equation}
\mathrm{MRA}=\frac{1}{10}\sum_{\theta}\mathbb{1}\!\left(\frac{|\hat{y}-y|}{y}<1-\theta\right).
\setlength\abovedisplayskip{4pt}%
\setlength\belowdisplayskip{4pt}%
\end{equation}
Because any question may be declined, the overall score decomposes as: 
% $S=(1-u)\,\mathrm{AR}\cdot\mathrm{Sel}+u\,\mathrm{Rej}$
\begin{equation}
S=(1-u)\,\mathrm{AR}\cdot\mathrm{Sel}+u\,\mathrm{Rej},
\setlength\abovedisplayskip{4pt}%
\setlength\belowdisplayskip{4pt}%
\label{eq:score}
\end{equation}
where $u$ is the share of unanswerable questions, the answer rate AR is the fraction of answerable questions a model attempts, the selective score Sel is its score on those attempts, and Rej is the rate of correct refusals. AR reflects whether the evidence was found and Sel whether it was understood. Declining an answerable question earns nothing, so $S$ rewards guessing, and we therefore report Rej alongside it.

\section{Experiments and Analysis}
\label{sec:experiments}

\subsection{Experimental Setup}
\label{sec:setup}

\paragraph{Models.}
We evaluate 23 models in three groups. The proprietary group spans two families at different strengths, Claude Opus, Sonnet, and Haiku 5.5, together with GPT-6.1 Sol and Terra and GPT-6 Luna. The open-weight group covers general-purpose models at multiple scales, including Qwen3.5~\cite{qwen3.5}, Qwen3.6~\cite{qwen3.6-27b,qwen3.6-35b-a3b}, Qwen3.8~\cite{qwen3.8}, Gemma-4~\cite{team2026gemma}, InternVL3.5~\cite{wang2025internvl3}, LLaVA-OneVision~\cite{li2024llava}, GLM-4.6V~\cite{hong2025glm}, and Kimi-VL~\cite{team2025kimi}, all served with vLLM~\cite{kwon2023efficientmemorymanagementlarge}. The spatially specialized group consists of SenseNova-SI~\cite{cai2026scaling}, VST~\cite{yang2026visual}, Cambrian-S~\cite{yang2026cambrian}, RoboBrain~2.0~\cite{team2025robobrain}, and Cosmos-Reason1~\cite{azzolini2025cosmos}, which are trained on dedicated spatial or embodied reasoning data and thus probe whether gains from such supervision transfer to active exploration.

\vspace{-1em}
\paragraph{Implementation.}
All models receive identical prompts and images at $640{\times}480$ resolution. Proprietary models run at each provider's high reasoning effort, and output budgets including reasoning tokens are 32{,}768 tokens, with truncated responses counted as unanswered. We report results on the 227 episodes whose shortest evidence route fits within the step budget, and the remaining in Appendix~\ref{app:hyper}.
\vspace{-1em}
\paragraph{Baselines.}
Random Guessing gives the expected score of answering uniformly at random. As a human reference, expert annotators answered questions from the oracle views without pose information, confirming that the questions are solvable given the right evidence. Full details see Appendix.

\subsection{Main Results}
\label{sec:main}

\myfinding{Passive performance does not predict active performance}
The active setting reorders models that the reference settings rank alike (Table~\ref{tab:main}). Among proprietary models, GPT-6.1 Sol and Terra lead with passive views and along the oracle route, yet Claude Opus leads once models must collect views themselves. Among open-weight models, Qwen3.8-27B is the best in every setting that provides images but trails two Qwen3.6 models in the active one. No spatially specialized model gains more than one point from the oracle route over Blind, so passive spatial training has not taught them to use evidence gathered along a trajectory.

\myfinding{Collecting evidence, not reasoning over it, limits the strongest models}
Along the oracle route, the best proprietary models match the human reference, which answers from oracle views. With their own exploration, however, only Claude Opus, Claude Sonnet, and GPT-6.1 Sol explore and still exceed a random walk of the same budget, and their loss relative to the oracle route is three to ten times this gain. The other three proprietary models fall below the random walk, including GPT-6.1 Terra, which is among the best with passive views. Cosmos-Reason1 also shows a positive gap, but it answers without moving and scores below its own Blind result. Most open-weight and specialized models likewise collect less useful evidence than an undirected walk of equal length. The remaining headroom for current VLMs therefore lies mainly in how they explore rather than in how they reason over given views.

\begin{figure*}[!t]
    \centering
    \includegraphics[width=\linewidth]{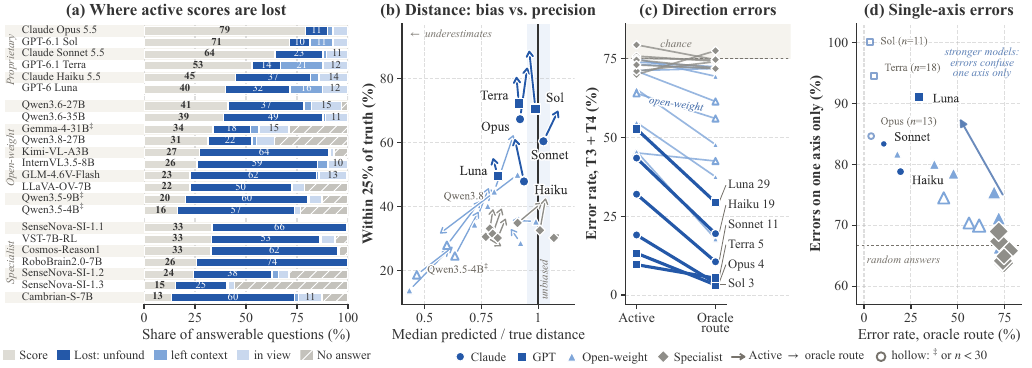}
    \vspace{-2em}
\caption{\textbf{Where scores are lost and how answers go wrong.} (a) Answerable questions in the active setting, split into score earned, score lost by where the evidence was at answer time, and no answer. (b) Distance bias against precision (share within 25\% of truth). (c) Error rate on directional questions (T3\,+\,T4). (d) Share of direction errors on one axis only against the oracle-route error rate. Arrows and lines in (b, c) run from the active setting to the oracle route. Hollow markers: responses affected by truncation ($^{\ddagger}$), or fewer than 30 errors in (d).}
    \label{fig:errors}
    \vspace{-1em}
\end{figure*}

\subsection{Do VLMs Explore Purposefully?}
\label{sec:behavior}
% \vspace{-0.8em}
\myfinding{Few models stop at the right time}
Claude models stop early after a median of 11 to 17 steps and never exhaust the budget, although random-walk coverage still rises from 36.7\% at 20 steps to 56.2\% at 40 (Fig.~\ref{fig:behavior}a,\,b). GPT models instead walk until forced, yet about a third of Sol's and Terra's moves return to visited views, and Luna spends 38 steps on only 16 distinct views. At the other extreme, Kimi-VL, Cosmos-Reason1, SenseNova-SI-1.1, and RoboBrain~2.0 mostly answer from the start view.
 \begin{table}[t]
\centering
\footnotesize
\caption{\textbf{Exploration gain under matched memory} (\%). RW-8 replays the RW-40 route but shows only the start image and the last eight views, as in the active setting. $\Delta_{\mathrm{RW}}^{\mathrm{mem}}$ = Active $-$ RW-8.}
\vspace{-0.8em}
\setlength{\tabcolsep}{4.2pt}
\renewcommand{\arraystretch}{1.04}
\begin{tabular}{l ccc cc}
\toprule
\textbf{Model} & Active & RW-40 & RW-8 & $\Delta_{\mathrm{RW}}$ & $\Delta_{\mathrm{RW}}^{\mathrm{mem}}$ \\
\midrule
GPT-6.1 Sol       & 68.7 & 61.8 & 43.2 & $+6.9$   & $\mathbf{+25.6}$ \\
GPT-6.1 Terra     & 55.1 & 61.4 & 37.8 & $-6.3$       & $\mathbf{+17.3}$ \\
Claude Sonnet 5.5 & 61.2 & 59.2 & 46.8 & $+2.0$       & $\mathbf{+14.4}$ \\
Claude Opus 5.5   & 74.0 & 70.1 & 61.7 & $+3.9$   & $\mathbf{+12.3}$ \\
Claude Haiku 5.5  & 45.2 & 55.6 & 43.2 & $-10.4$  & $+2.0$ \\
GPT-6 Luna        & 40.3 & 54.0 & 46.5 & $-13.7$  & $-6.2$ \\
\bottomrule
\end{tabular}
\vspace{-2em}
\label{tab:memory}
\end{table}
\myfinding{Strong models seek evidence with purpose, while most others find less than a random walk}
We compare each model with a length-matched random walk (Fig.~\ref{fig:behavior}c). Every proprietary model except GPT-6 Luna observes all evidence significantly more often than this walk, by up to 33.5 points, whether it stops early like the Claude models or uses the whole budget like Sol and Terra. Luna walks nearly as far as Sol yet finds no more, so purposeful search does not depend on exploration length. Of the 17 open-weight and specialist models, 12 fall significantly below the walk, by up to 43 points, as they shuttle among a few views or rotate in place and cover less of a scene than random steps.
\begin{figure*}[!t]
    \centering
    \includegraphics[width=\linewidth]{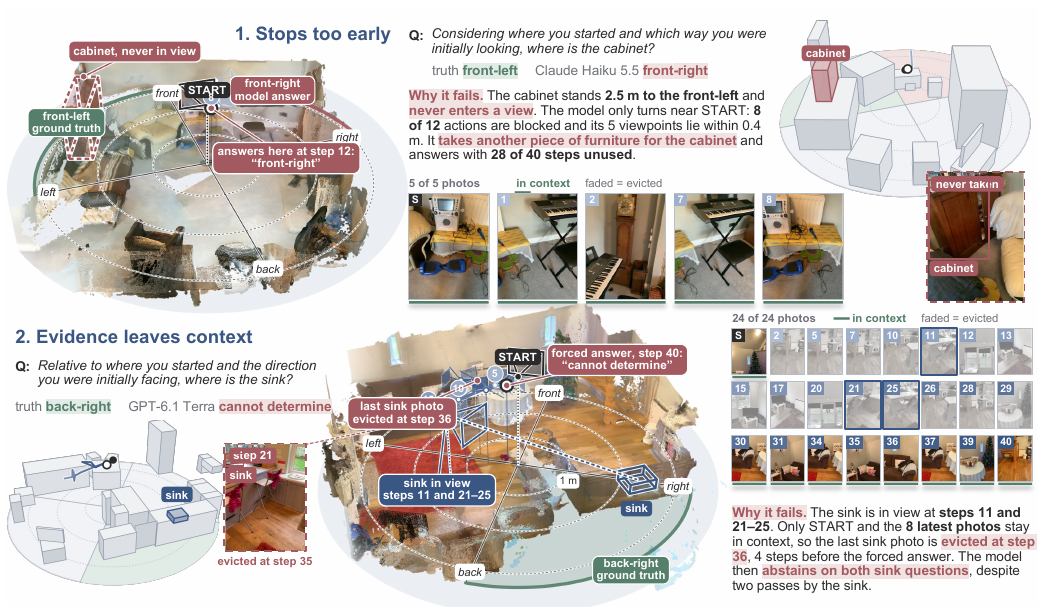}
    \vspace{-2em}
    \caption{\textbf{Representative failures}: stopping too early and evidence leaving context. More failure cases are displayed in Appendix~\ref{app:case_study}.}
    \label{fig:failures}
    \vspace{-1em}
\end{figure*} 
\myfinding{Limited memory caps the exploration gain of strong models}
Finding more evidence than the random walk yields only a small $\Delta_{\mathrm{RW}}$, because the walk answers with every view in context while the active agent keeps only recent ones. With the walk limited to the same window (Table~\ref{tab:memory}), the four strongest models gain between $+12.3$ and $+25.6$, with Terra reversing from $-6.3$ to $+17.3$, whereas Claude Haiku and GPT-6 Luna stay at or below zero.

\subsection{Do VLMs Know When They Don't Know?}
\label{sec:abstention}

\myfinding{Few models abstain according to the evidence}
Fig.~\ref{fig:behavior}(d) compares how often models decline answerable questions with the evidence in view and never found; a calibrated model would decline only in the latter case. GPT-6.1 Terra comes closest and also rejects 84.1\% of unanswerable questions. GPT-6.1 Sol, GPT-6 Luna, and Claude Haiku decline only 32 to 42\% of questions whose evidence they never found, and Claude Opus and Sonnet almost never abstain. Several open-weight and specialist models do not separate the two cases: GLM-4.6V declines about 30\% either way, while Cambrian-S and Kimi-VL decline more often with the evidence in view. A high rejection rate alone therefore does not show that a model knows what it has not seen.

\myfinding{Models guess knowingly, and the total score rewards it}
In 37.4\% of its episodes, Opus notes that an object was never seen and then answers anyway, e.g., ``table never seen, so Q2 and Q3 are guesses''. Since declining an answerable question earns nothing, this strategy is rational under the total score, and Opus obtains the highest Active score with a rejection rate of zero. Even along the oracle route, where only the masked target is unobserved, it rejects just 9 of the 63 unanswerable questions. Willingness to abstain must therefore be reported apart from the score.

\subsection{Finding or Reasoning? Locating the Bottleneck}
\label{sec:bottleneck}

\myfinding{Strong models lose score to finding and remembering evidence}
Fig.~\ref{fig:errors}(a) splits the score each model loses by the state of the evidence when it answers. Models that stop early, such as Claude Haiku and Sonnet, lose mainly to evidence they never found, whereas GPT-6.1 Terra, which exhausts the budget, loses most to evidence that has left its context. Models also differ in how much survives eviction: Claude Opus drops only from 0.872 to 0.832 once the evidence is out of context, while Terra falls from 0.776 to 0.258. Terra does not misremember: it declines 71.1\% of these questions and stays accurate on the rest (0.893), and a cognitive-map probe shows that it localizes 91.4\% of the seen objects still in context but only 14.6\% of those evicted.

\myfinding{Weak models cannot use the evidence even when it is in view}
With all evidence in context, strong proprietary models score 0.78 to 0.88 and the larger open-weight models 0.50 to 0.58. InternVL3.5, LLaVA-OV, GLM-4.6V, Kimi-VL, and the specialists score only 0.13 to 0.38, little above their 0.13 to 0.33 with the evidence never seen. Unfound evidence is their largest loss in Fig.~\ref{fig:errors}(a), yet finding it would not help: the specialists, InternVL3.5, and LLaVA-OV gain at most one point from the oracle route over Blind (Table~\ref{tab:main}), so their bottleneck is reasoning rather than exploration.

\subsection{Error Analysis}
\label{sec:errors}

\myfinding{Distances are systematically underestimated}
The median ratio of predicted to true distance is below one for 20 of 23 models (Fig.~\ref{fig:errors}b). The oracle route makes strong models more precise but does not remove the bias: the share within 25\% of the truth rises from about 70\% to 84\% for Opus and 89\% for GPT-6.1 Sol, while their medians stay near 0.95. Since the bias survives with all evidence supplied, it stems from metric scale estimation, not missing views.

\myfinding{Direction errors are random for weak models and confined to one axis for capable ones}
Along the oracle route, the direction error rate falls to 4\% for Opus and 3\% for Sol (Fig.~\ref{fig:errors}c). For InternVL3.5, LLaVA-OV, GLM-4.6V, Kimi-VL, and the specialists, it stays at 69 to 79\% in both settings, near the 75\% of random answers, with errors split evenly among the three wrong quadrants (Fig.~\ref{fig:errors}d). In contrast, 91\% of GPT-6 Luna's and 83\% of Claude Sonnet's oracle-route errors are on one axis only: these models locate the object but misalign the reference frame. The same confusion appears in linked questions: proprietary models violate at most 2.9\% of distance-comparison constraints but 10 to 37\% of those between mirrored perspective questions.

Fig.~\ref{fig:failures} shows two representative failures: stopping too early and evidence leaving the context. Refer to Appendix~\ref{app:case_study} for cases answering wrongly despite the evidence shown in the view. Each points to a remedy: stopping rules tied to the evidence collected rather than the budget, spatial memory that persists beyond the image context, and more reliable transformation between reference frames.

% \vspace{-0.4em}
\section{Conclusion}
\label{sec:conclusion}

We presented \textsc{ActSpace-Bench}, which evaluates spatial reasoning when a model must find its own evidence in real indoor scenes. Comparing its exploration with a random walk and an oracle route separates finding evidence from reasoning. Across 23 VLMs, the strongest seek evidence with purpose but lose much of it once it leaves context, whereas weaker models explore no better than random walk or cannot use supplied evidence. This calls for evidence-based stopping rules, spatial memory beyond the image context, and reliable reference-frame transformation. 

{
    \small
    \bibliographystyle{ieeenat_fullname}
    \bibliography{main}

\begin{thebibliography}{43}
\providecommand{\natexlab}[1]{#1}
\providecommand{\url}[1]{\texttt{#1}}
\expandafter\ifx\csname urlstyle\endcsname\relax
  \providecommand{\doi}[1]{doi: #1}\else
  \providecommand{\doi}{doi: \begingroup \urlstyle{rm}\Url}\fi

\bibitem[Azzolini et~al.(2025)Azzolini, Bai, Brandon, Cao, Chattopadhyay, Chen,
  Chu, Cui, Diamond, Ding, et~al.]{azzolini2025cosmos}
Alisson Azzolini, Junjie Bai, Hannah Brandon, Jiaxin Cao, Prithvijit
  Chattopadhyay, Huayu Chen, Jinju Chu, Yin Cui, Jenna Diamond, Yifan Ding,
  et~al.
\newblock Cosmos-reason1: From physical common sense to embodied reasoning.
\newblock \emph{arXiv preprint arXiv:2503.15558}, 2025.

\bibitem[Baruch et~al.(2021)Baruch, Chen, Dehghan, Dimry, Feigin, Fu, Gebauer,
  Joffe, Kurz, Schwartz, et~al.]{baruch2021arkitscenes}
Gilad Baruch, Zhuoyuan Chen, Afshin Dehghan, Tal Dimry, Yuri Feigin, Peter Fu,
  Thomas Gebauer, Brandon Joffe, Daniel Kurz, Arik Schwartz, et~al.
\newblock Arkitscenes: A diverse real-world dataset for 3d indoor scene
  understanding using mobile rgb-d data.
\newblock \emph{arXiv preprint arXiv:2111.08897}, 2021.

\bibitem[Black et~al.(2024)Black, Brown, Driess, Esmail, Equi, Finn, Fusai,
  Groom, Hausman, Ichter, et~al.]{black2024pi_0}
Kevin Black, Noah Brown, Danny Driess, Adnan Esmail, Michael Equi, Chelsea
  Finn, Niccolo Fusai, Lachy Groom, Karol Hausman, Brian Ichter, et~al.
\newblock ${\pi_0}$: A vision-language-action flow model for general robot
  control.
\newblock \emph{arXiv preprint arXiv:2410.24164}, 2024.

\bibitem[Cai et~al.(2025)Cai, Wang, Sun, Wang, Gu, Yin, Lin, Yang, Wei, Qian,
  et~al.]{cai2025holistic}
Zhongang Cai, Yubo Wang, Qingping Sun, Ruisi Wang, Chenyang Gu, Wanqi Yin,
  Zhiqian Lin, Zhitao Yang, Chen Wei, Oscar Qian, et~al.
\newblock Holistic evaluation of multimodal llms on spatial intelligence.
\newblock \emph{arXiv preprint arXiv:2508.13142}, 2025.

\bibitem[Cai et~al.(2026)Cai, Wang, Gu, Pu, Xu, Wang, Yin, Yang, Wei, Zhou,
  et~al.]{cai2026scaling}
Zhongang Cai, Ruisi Wang, Chenyang Gu, Fanyi Pu, Junxiang Xu, Yubo Wang, Wanqi
  Yin, Zhitao Yang, Chen Wei, Tongxi Zhou, et~al.
\newblock Scaling spatial intelligence with multimodal foundation models.
\newblock In \emph{Proceedings of the IEEE/CVF Conference on Computer Vision
  and Pattern Recognition}, pages 7879--7890, 2026.

\bibitem[Chen et~al.(2024)Chen, Xu, Kirmani, Ichter, Sadigh, Guibas, and
  Xia]{chen2024spatialvlm}
Boyuan Chen, Zhuo Xu, Sean Kirmani, Brain Ichter, Dorsa Sadigh, Leonidas
  Guibas, and Fei Xia.
\newblock Spatialvlm: Endowing vision-language models with spatial reasoning
  capabilities.
\newblock In \emph{Proceedings of the IEEE/CVF conference on computer vision
  and pattern recognition}, pages 14455--14465, 2024.

\bibitem[Cheng et~al.(2024)Cheng, Yin, Fu, Guo, Yang, Kautz, Wang, and
  Liu]{cheng2024spatialrgpt}
An-Chieh Cheng, Hongxu Yin, Yang Fu, Qiushan Guo, Ruihan Yang, Jan Kautz,
  Xiaolong Wang, and Sifei Liu.
\newblock Spatialrgpt: Grounded spatial reasoning in vision-language models.
\newblock \emph{Advances in Neural Information Processing Systems},
  37:\penalty0 135062--135093, 2024.

\bibitem[Fan et~al.(2026)Fan, Zhang, Li, Zhang, Chen, Hu, Wang, Wang, Qu, Zhou,
  et~al.]{fan2026vlm}
Zhiwen Fan, Jian Zhang, Renjie Li, Junge Zhang, Runjin Chen, Hezhen Hu, Kevin
  Wang, Peihao Wang, Huaizhi Qu, Shijie Zhou, et~al.
\newblock Vlm-3r: Vision-language models augmented with instruction-aligned 3d
  reconstruction.
\newblock In \emph{Proceedings of the IEEE/CVF conference on computer vision
  and pattern recognition}, pages 31054--31065, 2026.

\bibitem[Hong et~al.(2025)Hong, Yu, Gu, Wang, Gan, Tang, Cheng, Qi, Ji, Pan,
  et~al.]{hong2025glm}
Wenyi Hong, Wenmeng Yu, Xiaotao Gu, Guo Wang, Guobing Gan, Haomiao Tang, Jiale
  Cheng, Ji Qi, Junhui Ji, Lihang Pan, et~al.
\newblock Glm-4.5 v and glm-4.1 v-thinking: Towards versatile multimodal
  reasoning with scalable reinforcement learning.
\newblock \emph{arXiv preprint arXiv:2507.01006}, 2025.

\bibitem[Hong et~al.(2026)Hong, Liu, Yin, Li, Guibas, Fei-Fei, Wu, and
  Choi]{hong2026esibenchembodiedspatialintelligence}
Yining Hong, Jiageng Liu, Han Yin, Manling Li, Leonidas Guibas, Li Fei-Fei,
  Jiajun Wu, and Yejin Choi.
\newblock Esi-bench: Towards embodied spatial intelligence that closes the
  perception-action loop, 2026.

\bibitem[Jiang et~al.(2025)Jiang, Liu, Chen, Luo, Chen, Pan, Li, and
  Lin]{jiang2025beyond}
Kaixuan Jiang, Yang Liu, Weixing Chen, Jingzhou Luo, Ziliang Chen, Ling Pan,
  Guanbin Li, and Liang Lin.
\newblock Beyond the destination: A novel benchmark for exploration-aware
  embodied question answering.
\newblock In \emph{2025 IEEE/CVF International Conference on Computer Vision
  (ICCV)}, pages 9091--9101. IEEE, 2025.

\bibitem[Kim et~al.(2024)Kim, Pertsch, Karamcheti, Xiao, Balakrishna, Nair,
  Rafailov, Foster, Lam, Sanketi, et~al.]{kim2024openvla}
Moo~Jin Kim, Karl Pertsch, Siddharth Karamcheti, Ted Xiao, Ashwin Balakrishna,
  Suraj Nair, Rafael Rafailov, Ethan Foster, Grace Lam, Pannag Sanketi, et~al.
\newblock Openvla: An open-source vision-language-action model.
\newblock \emph{arXiv preprint arXiv:2406.09246}, 2024.

\bibitem[Kwon et~al.(2023)Kwon, Li, Zhuang, Sheng, Zheng, Yu, Gonzalez, Zhang,
  and Stoica]{kwon2023efficientmemorymanagementlarge}
Woosuk Kwon, Zhuohan Li, Siyuan Zhuang, Ying Sheng, Lianmin Zheng, Cody~Hao Yu,
  Joseph Gonzalez, Hao Zhang, and Ion Stoica.
\newblock Efficient memory management for large language model serving with
  pagedattention.
\newblock In \emph{Proceedings of the 29th symposium on operating systems
  principles}, pages 611--626, 2023.

\bibitem[Li et~al.(2024)Li, Zhang, Guo, Zhang, Li, Zhang, Zhang, Zhang, Li,
  Liu, et~al.]{li2024llava}
Bo Li, Yuanhan Zhang, Dong Guo, Renrui Zhang, Feng Li, Hao Zhang, Kaichen
  Zhang, Peiyuan Zhang, Yanwei Li, Ziwei Liu, et~al.
\newblock Llava-onevision: Easy visual task transfer.
\newblock \emph{arXiv preprint arXiv:2408.03326}, 2024.

\bibitem[Li et~al.(2025)Li, Li, Wang, Yan, Zhang, Chen, Hou, Jiang, Zhang,
  Shen, Lu, and
  Zhuang]{li2025viewspatialbenchevaluatingmultiperspectivespatial}
Dingming Li, Hongxing Li, Zixuan Wang, Yuchen Yan, Hang Zhang, Siqi Chen,
  Guiyang Hou, Shengpei Jiang, Wenqi Zhang, Yongliang Shen, Weiming Lu, and
  Yueting Zhuang.
\newblock Viewspatial-bench: Evaluating multi-perspective spatial localization
  in vision-language models, 2025.

\bibitem[Lin et~al.(2025)Lin, Xu, Zhu, Yang, Cao, Ran, Hu, Zhu, Xie, Long,
  et~al.]{lin2025mmsi}
Jingli Lin, Runsen Xu, Shaohao Zhu, Sihan Yang, Peizhou Cao, Yunlong Ran, Miao
  Hu, Chenming Zhu, Yiman Xie, Yilin Long, et~al.
\newblock Mmsi-video-bench: A holistic benchmark for video-based spatial
  intelligence.
\newblock \emph{arXiv preprint arXiv:2512.10863}, 2025.

\bibitem[Lin et~al.(2026)Lin, Zhu, Xu, Mao, Liu, Wang, and Pang]{lin2026ost}
Jingli Lin, Chenming Zhu, Runsen Xu, Xiaohan Mao, Xihui Liu, Tai Wang, and
  Jiangmiao Pang.
\newblock Ost-bench: Evaluating the capabilities of mllms in online
  spatio-temporal scene understanding.
\newblock \emph{Advances in Neural Information Processing Systems}, 38, 2026.

\bibitem[Liu et~al.(2025)Liu, Chen, Bai, Liang, Li, Gao, and
  Lin]{liu2025aligning}
Yang Liu, Weixing Chen, Yongjie Bai, Xiaodan Liang, Guanbin Li, Wen Gao, and
  Liang Lin.
\newblock Aligning cyber space with physical world: A comprehensive survey on
  embodied ai.
\newblock \emph{IEEE/ASME Transactions on Mechatronics}, 2025.

\bibitem[Majumdar et~al.(2024)Majumdar, Ajay, Zhang, Putta, Yenamandra, Henaff,
  Silwal, Mcvay, Maksymets, Arnaud, et~al.]{majumdar2024openeqa}
Arjun Majumdar, Anurag Ajay, Xiaohan Zhang, Pranav Putta, Sriram Yenamandra,
  Mikael Henaff, Sneha Silwal, Paul Mcvay, Oleksandr Maksymets, Sergio Arnaud,
  et~al.
\newblock Openeqa: Embodied question answering in the era of foundation models.
\newblock In \emph{2024 IEEE/CVF Conference on Computer Vision and Pattern
  Recognition (CVPR)}, pages 16488--16498. IEEE, 2024.

\bibitem[{Qwen Team}(2026{\natexlab{a}})]{qwen3.5}
{Qwen Team}.
\newblock {Qwen3.5}: Towards native multimodal agents, 2026{\natexlab{a}}.

\bibitem[{Qwen Team}(2026{\natexlab{b}})]{qwen3.6-27b}
{Qwen Team}.
\newblock {Qwen3.6-27B}: Flagship-level coding in a {27B} dense model,
  2026{\natexlab{b}}.

\bibitem[{Qwen Team}(2026{\natexlab{c}})]{qwen3.6-35b-a3b}
{Qwen Team}.
\newblock {Qwen3.6-35B-A3B}: Agentic coding power, now open to all,
  2026{\natexlab{c}}.

\bibitem[{Qwen Team}(2026{\natexlab{d}})]{qwen3.8}
{Qwen Team}.
\newblock {Qwen3.8-Max}: A new bar for coding and cowork, 2026{\natexlab{d}}.

\bibitem[Ren et~al.(2024)Ren, Clark, Dixit, Itkina, Majumdar, and
  Sadigh]{ren2024explore}
Allen~Z Ren, Jaden Clark, Anushri Dixit, Masha Itkina, Anirudha Majumdar, and
  Dorsa Sadigh.
\newblock Explore until confident: Efficient exploration for embodied question
  answering.
\newblock \emph{arXiv preprint arXiv:2403.15941}, 2024.

\bibitem[Sakamoto et~al.(2026)Sakamoto, Miyanishi, Azuma, Kurita, Morikuni,
  Chiba, Kawanabe, Iwasawa, and Matsuo]{sakamoto2026e3vs}
Koya Sakamoto, Taiki Miyanishi, Daichi Azuma, Shuhei Kurita, Shu Morikuni,
  Naoya Chiba, Motoaki Kawanabe, Yusuke Iwasawa, and Yutaka Matsuo.
\newblock E3vs-bench: A benchmark for viewpoint-dependent active perception in
  3d gaussian splatting scenes.
\newblock In \emph{European Conference on Computer Vision}, pages 60--78.
  Springer, 2026.

\bibitem[Team et~al.(2025{\natexlab{a}})Team, Cao, Tan, Ji, Chen, Lin, Li, Cao,
  Wang, Zhou, et~al.]{team2025robobrain}
BAAI~RoboBrain Team, Mingyu Cao, Huajie Tan, Yuheng Ji, Xiansheng Chen, Minglan
  Lin, Zhiyu Li, Zhou Cao, Pengwei Wang, Enshen Zhou, et~al.
\newblock Robobrain 2.0 technical report.
\newblock \emph{arXiv preprint arXiv:2507.02029}, 2025{\natexlab{a}}.

\bibitem[Team et~al.(2026)Team, Abd, Aggarwal, Algayres, Andreev, Bachem,
  Ballantyne, Brick, C{\u{a}}rbune, Casbon, et~al.]{team2026gemma}
Gemma Team, Sherif~El Abd, Vaibhav Aggarwal, Robin Algayres, Alek Andreev,
  Olivier Bachem, Ian Ballantyne, Cormac Brick, Victor C{\u{a}}rbune, Michelle
  Casbon, et~al.
\newblock Gemma 4 technical report.
\newblock \emph{arXiv preprint arXiv:2607.02770}, 2026.

\bibitem[Team et~al.(2025{\natexlab{b}})Team, Abeyruwan, Ainslie, Alayrac,
  Arenas, Armstrong, Balakrishna, Baruch, Bauza, Blokzijl,
  et~al.]{team2025gemini}
Gemini~Robotics Team, Saminda Abeyruwan, Joshua Ainslie, Jean-Baptiste Alayrac,
  Montserrat~Gonzalez Arenas, Travis Armstrong, Ashwin Balakrishna, Robert
  Baruch, Maria Bauza, Michiel Blokzijl, et~al.
\newblock Gemini robotics: Bringing ai into the physical world.
\newblock \emph{arXiv preprint arXiv:2503.20020}, 2025{\natexlab{b}}.

\bibitem[Team et~al.(2025{\natexlab{c}})Team, Du, Yin, Xing, Qu, Wang, Chen,
  Zhang, Du, Wei, et~al.]{team2025kimi}
Kimi Team, Angang Du, Bohong Yin, Bowei Xing, Bowen Qu, Bowen Wang, Cheng Chen,
  Chenlin Zhang, Chenzhuang Du, Chu Wei, et~al.
\newblock Kimi-vl technical report.
\newblock \emph{arXiv preprint arXiv:2504.07491}, 2025{\natexlab{c}}.

\bibitem[Wang et~al.(2026)Wang, Yin, Zhang, Zhang, Wang, Wang, Zhang,
  Chandrasegaran, Liu, Krishna, Xie, Wu, Fei-Fei, and
  Li]{wang2026mindcubespatialmentalmodeling}
Qineng Wang, Baiqiao Yin, Pingyue Zhang, Jianshu Zhang, Kangrui Wang, Zihan
  Wang, Jieyu Zhang, Keshigeyan Chandrasegaran, Han Liu, Ranjay Krishna,
  Saining Xie, Jiajun Wu, Li Fei-Fei, and Manling Li.
\newblock Mindcube: Spatial mental modeling from limited views, 2026.

\bibitem[Wang et~al.(2025{\natexlab{a}})Wang, Gao, Gu, Pu, Cui, Wei, Liu, Jing,
  Ye, Shao, et~al.]{wang2025internvl3}
Weiyun Wang, Zhangwei Gao, Lixin Gu, Hengjun Pu, Long Cui, Xingguang Wei,
  Zhaoyang Liu, Linglin Jing, Shenglong Ye, Jie Shao, et~al.
\newblock Internvl3. 5: Advancing open-source multimodal models in versatility,
  reasoning, and efficiency.
\newblock \emph{arXiv preprint arXiv:2508.18265}, 2025{\natexlab{a}}.

\bibitem[Wang et~al.(2025{\natexlab{b}})Wang, Chen, Luo, Dong, Zhang, Xu, Wang,
  Li, and Liu]{wang2025actiview}
Ziyue Wang, Chi Chen, Fuwen Luo, Yurui Dong, Yuanchi Zhang, Yuzhuang Xu,
  Xiaolong Wang, Peng Li, and Yang Liu.
\newblock Actiview: Evaluating active perception ability for multimodal large
  language models.
\newblock In \emph{Proceedings of the 63rd Annual Meeting of the Association
  for Computational Linguistics (Volume 1: Long Papers)}, pages 7605--7633,
  2025{\natexlab{b}}.

\bibitem[Wu et~al.(2026)Wu, Liu, Hung, and Duan]{wu2026spatial}
Diankun Wu, Fangfu Liu, Yi-Hsin Hung, and Yueqi Duan.
\newblock Spatial-mllm: Boosting mllm capabilities in visual-based spatial
  intelligence.
\newblock \emph{Advances in neural information processing systems},
  38:\penalty0 13569--13597, 2026.

\bibitem[Wu and Xie(2024)]{wu2024v}
Penghao Wu and Saining Xie.
\newblock V*: Guided visual search as a core mechanism in multimodal llms.
\newblock In \emph{2024 IEEE/CVF Conference on Computer Vision and Pattern
  Recognition (CVPR)}, pages 13084--13094. IEEE, 2024.

\bibitem[Yang et~al.(2025{\natexlab{a}})Yang, Liu, Guo, Dong, Zhang, Zhang,
  Wang, Zhou, Xie, Wang, et~al.]{yang2025egolife}
Jingkang Yang, Shuai Liu, Hongming Guo, Yuhao Dong, Xiamengwei Zhang, Sicheng
  Zhang, Pengyun Wang, Zitang Zhou, Binzhu Xie, Ziyue Wang, et~al.
\newblock Egolife: Towards egocentric life assistant.
\newblock In \emph{2025 IEEE/CVF Conference on Computer Vision and Pattern
  Recognition (CVPR)}, pages 28885--28900. IEEE, 2025{\natexlab{a}}.

\bibitem[Yang et~al.(2025{\natexlab{b}})Yang, Yang, Gupta, Han, Fei-Fei, and
  Xie]{yang2025thinking}
Jihan Yang, Shusheng Yang, Anjali~W Gupta, Rilyn Han, Li Fei-Fei, and Saining
  Xie.
\newblock Thinking in space: How multimodal large language models see,
  remember, and recall spaces.
\newblock In \emph{2025 IEEE/CVF Conference on Computer Vision and Pattern
  Recognition (CVPR)}, pages 10632--10643. IEEE, 2025{\natexlab{b}}.

\bibitem[Yang et~al.(2026{\natexlab{a}})Yang, Zhu, Li, Huang, Yan, Zhou, Liu,
  Li, Li, Wang, et~al.]{yang2026visual}
Rui Yang, Ziyu Zhu, Yanwei Li, Jingjia Huang, Shen Yan, Siyuan Zhou, Zhe Liu,
  Xiangtai Li, Shuangye Li, Wenqian Wang, et~al.
\newblock Visual spatial tuning.
\newblock In \emph{European Conference on Computer Vision}, pages 192--211.
  Springer, 2026{\natexlab{a}}.

\bibitem[Yang et~al.(2026{\natexlab{b}})Yang, Xu, Xie, Yang, Li, Lin, Zhu,
  Chen, Duan, Yue, Lin, Wang, and
  Pang]{yang2026mmsibenchbenchmarkmultiimagespatial}
Sihan Yang, Runsen Xu, Yiman Xie, Sizhe Yang, Mo Li, Jingli Lin, Chenming Zhu,
  Xiaochen Chen, Haodong Duan, Xiangyu Yue, Dahua Lin, Tai Wang, and Jiangmiao
  Pang.
\newblock Mmsi-bench: A benchmark for multi-image spatial intelligence,
  2026{\natexlab{b}}.

\bibitem[Yang et~al.(2026{\natexlab{c}})Yang, Yang, Huang, Brown, Yang, Yu,
  Tong, Zheng, Xu, Wang, et~al.]{yang2026cambrian}
Shusheng Yang, Jihan Yang, Pinzhi Huang, Ellis Brown, Zihao Yang, Yue Yu,
  Shengbang Tong, Zihan Zheng, Yifan Xu, Muhan Wang, et~al.
\newblock Cambrian-s: Towards spatial supersensing in video.
\newblock In \emph{International Conference on Learning Representations}, pages
  78185--78225, 2026{\natexlab{c}}.

\bibitem[Yang et~al.(2025{\natexlab{c}})Yang, Yang, Zhou, Chen, Zhang, Du, and
  Gan]{yang20253d}
Yuncong Yang, Han Yang, Jiachen Zhou, Peihao Chen, Hongxin Zhang, Yilun Du, and
  Chuang Gan.
\newblock 3d-mem: 3d scene memory for embodied exploration and reasoning.
\newblock In \emph{2025 IEEE/CVF Conference on CVPR}, pages 17294--17303. IEEE,
  2025{\natexlab{c}}.

\bibitem[Yeh et~al.(2026)Yeh, Wang, Tong, Cheng, Wang, Chu, Zhai, Chen, Gao,
  and Ma]{yeh2026seeing}
Chun-Hsiao Yeh, Chenyu Wang, Shengbang Tong, Ta-Ying Cheng, Ruoyu Wang, Tianzhe
  Chu, Yuexiang Zhai, Yubei Chen, Shenghua Gao, and Yi Ma.
\newblock Seeing from another perspective: Evaluating multi-view understanding
  in mllms.
\newblock In \emph{Proceedings of the AAAI Conference on Artificial
  Intelligence}, pages 12000--12008, 2026.

\bibitem[Yeshwanth et~al.(2023)Yeshwanth, Liu, Nie{\ss}ner, and
  Dai]{yeshwanth2023scannet++}
Chandan Yeshwanth, Yueh-Cheng Liu, Matthias Nie{\ss}ner, and Angela Dai.
\newblock Scannet++: A high-fidelity dataset of 3d indoor scenes.
\newblock In \emph{2023 IEEE/CVF ICCV}, pages 12--22. IEEE, 2023.

\bibitem[Zhang et~al.(2025)Zhang, Wang, Zhang, Dai, Xia, Yan, Hong, and
  Zhao]{zhang2025dsi}
Ziang Zhang, Zehan Wang, Guanghao Zhang, Weilong Dai, Yan Xia, Ziang Yan,
  Minjie Hong, and Zhou Zhao.
\newblock Dsi-bench: A benchmark for dynamic spatial intelligence.
\newblock \emph{arXiv preprint arXiv:2510.18873}, 2025.

\end{thebibliography}
}
\appendix
\clearpage
\setcounter{page}{1}
\maketitlesupplementary

% ===== Appendix table style (shared by all appendix tables) =====
\definecolor{apHead}{HTML}{E3E9F1}    % header band, muted blue-grey
\definecolor{apStripe}{HTML}{F5F6F8}  % light neutral tint
\definecolor{apFlag}{HTML}{F8E7E1}    % muted red tint for flagged cells/rows
\definecolor{apRule}{gray}{0.62}      % thin rule inside a group
\newcolumntype{Y}{>{\centering\arraybackslash}X}
\newcommand{\apsetup}{\footnotesize\renewcommand{\arraystretch}{1.12}\setlength{\aboverulesep}{0pt}\setlength{\belowrulesep}{0pt}\setlength{\abovecaptionskip}{0pt}\setlength{\belowcaptionskip}{4pt}}
\newcommand{\aphead}{\rowcolor{apHead}}
\newcommand{\apgroup}[2]{\rowcolor{asGroup}\multicolumn{#1}{@{\hspace{3pt}}l}{\textit{\color{black!75}#2}}\\}
\newcommand{\apsep}{\arrayrulecolor{apRule}\specialrule{0.3pt}{0pt}{0pt}\arrayrulecolor{black}}
\newcommand{\apflag}[1]{\cellcolor{apFlag}\textbf{#1}}
\newcommand{\apzero}{\textcolor{asMuted}{0.0}}

\section{Benchmark Details}
\label{app:benchmark}

\subsection{Construction Details}
\label{app:construction}

This section gives the rules and thresholds behind each step of Sec.~\ref{sec:pipeline}. All thresholds are fixed in one configuration file, and every random choice is seeded per scene, so the entire benchmark can be rebuilt exactly from scratch.
\vspace{-1em}
\paragraph{Scene Parsing.}
We use the validation split of the 3D object detection subset of ARKitScenes~\cite{baruch2021arkitscenes}, keeping one capture per physical location, and the validation scenes of ScanNet++~\cite{yeshwanth2023scannet++}. All scenes are converted to a common metric frame with the $z$ axis up, and images are rotated upright with intrinsics and poses updated. Each camera pose is validated by comparing mesh-rendered depth with sensor depth, and a pose is rejected when the median absolute error reaches 0.10\,m. A scene is discarded if more than 20\% of its poses fail or fewer than 40 keyframes remain. Keyframes are selected when the camera has moved 0.30\,m or turned $15^{\circ}$ since the last one, taking the sharpest frame within a 0.5\,s window. An object can be asked about only if its category occurs exactly once in the scene, it has at least 100 mesh points, and its longest side is at least 0.10\,m.
\vspace{-1em}
\paragraph{View Graph.}
Keyframes are the nodes of a directed graph whose edges implement the six actions. For a node $i$ and a candidate $j$, let $d$ be the horizontal displacement, $\phi$ its bearing relative to the heading of $i$, and $\Delta\psi$ the change in heading. Table~\ref{tab:actions} lists the conditions. Only the largest strongly connected component is kept, so every view remains reachable from every other, and scenes that retain less than 70\% of their keyframes are discarded. A translation covers about 0.3\,m and a turn about $45^{\circ}$ at the median, and an action simply fails when no keyframe satisfies its condition.

\begin{table}[!htb]
\centering
\caption{\textbf{Action definitions.} Translations additionally require $0.2\,\text{m}\le|d|\le 1.5\,\text{m}$ and $|\Delta\psi|\le 45^{\circ}$; turns require $|d|\le 0.5\,\text{m}$.}
\label{tab:actions}
\apsetup
\begin{tabularx}{\columnwidth}{@{\hspace{4pt}}l l X@{\hspace{4pt}}}
\toprule
\aphead \textbf{Action} & \textbf{Condition on candidate $j$} & \textbf{Choice} \\
\midrule
forward     & $|\phi|\le 45^{\circ}$               & nearest \\
backward    & $|\phi|\ge 135^{\circ}$              & nearest \\
left        & $45^{\circ}<\phi<135^{\circ}$        & nearest \\
right       & $-135^{\circ}<\phi<-45^{\circ}$      & nearest \\
\apsep
\rowcolor{apStripe} turn left   & $20^{\circ}\le\Delta\psi\le 90^{\circ}$   & $\Delta\psi$ closest to $45^{\circ}$ \\
\rowcolor{apStripe} turn right  & $-90^{\circ}\le\Delta\psi\le -20^{\circ}$ & $\Delta\psi$ closest to $-45^{\circ}$ \\
\bottomrule
\end{tabularx}
\end{table}
% \vspace{-1em}
\paragraph{Visibility Labeling.}
For each object and view we sample up to 2{,}000 points of the object and project them into the image. A point counts as visible if it is occluded neither by the mesh, tested by ray casting with a 0.05\,m tolerance, nor by the sensor depth map, with a 0.10\,m tolerance. The second test corrects for holes in the mesh and for changes in the scene during capture. An object is \emph{visible} in a view when at least 30\% of its in-frustum points and 30\% of all its points are visible, at least 50 points are visible, their convex hull covers at least 2\% of the image, and at least 60\% of its projected box lies inside the frame. It is \emph{invisible} when fewer than 5\% of its in-frustum points are visible, and \emph{partial} otherwise. ScanNet++ applies slightly stricter conditions on the visible ratio and on image sharpness. On 50 randomly sampled visible pairs per source, at least 95\% were judged as being clearly recognizable by a human annotator.
\vspace{-1em}
\paragraph{Episode Sampling.}
An episode is built around a cluster of three to five askable objects that are pairwise at least 0.5\,m apart and never all visible in a single view. The start view is drawn at random from the views in which at least 70\% of the cluster is invisible while some annotated object can still be seen, so that the model starts with a meaningful image but without the evidence. Each episode contains three to five questions of at least two different task types.
\vspace{-1em}
\paragraph{Question Generation.}
Answers are computed from the centers of gravity-aligned object boxes, and a question is kept only if its answer is far from every decision boundary. Distances for T1 lie between 0.5 and 8\,m. For T2 and T5, the best and second-best candidates differ by at least 15\% and 0.3\,m. For T3, T4, and T5, the target direction is at least $20^{\circ}$ from every quadrant boundary, and the standpoint is at least 0.5\,m from the other objects on the horizontal plane, which removes objects stacked on top of each other. We further require that the evidence objects of a question can be related through views in which two of them appear together, so that their relative layout is inferable. For every question we compute by exhaustive search the minimum number of views that reveal all its evidence and the shortest action sequence that visits them, which defines the oracle route. Unanswerable questions are produced only by masking. All views in which the target object is visible or partial are removed, provided that they amount to at most 30\% of the graph, the remaining graph stays strongly connected, and the other questions of the episode remain answerable. Linked questions take two forms: two absolute distances paired with their comparison, and two perspective questions with the standpoint and facing object exchanged.
% \vspace{-1em}
\paragraph{Quality Control.}
Each template has 12 paraphrases, 84 in total, which were checked automatically to preserve placeholders and introduce no number, object name, or directional cue, and reviewed by the authors. To remove textual shortcuts, a random forest trained only on question text, option text, and template identity predicts the answers under five-fold cross-validation grouped by scene. Questions with the highest shortcut score are removed iteratively until no task type exceeds chance by over five points. Option letters are balanced within each source, and the four quadrants of T3 and T4 are balanced within five points of 25\%.

\subsection{Dataset Statistics}
\label{app:stats}

Table~\ref{tab:stats} summarizes the benchmark. The 994 questions comprise 248 on absolute distance (T1), 279 on relative distance (T2), 170 on perspective taking (T3), 209 on start-relative direction (T4), and 88 multi-hop questions (T5). The T2 questions ask for the closest object in 96 cases and the farthest in 98, and 85 are pairwise comparisons that appear only inside linked questions. The 64 unanswerable questions are 36 of type T1 and 28 of type T4. There are 141 constraints between linked questions, 85 between two distances and their comparison and 56 between mirrored perspective questions, and 93 episodes carry none.

\begin{table}[!htb]
\centering
\caption{\textbf{Benchmark statistics by source.}}
\label{tab:stats}
\apsetup
\begin{tabularx}{\columnwidth}{@{\hspace{4pt}}l YY >{\columncolor{apStripe}}Y@{\hspace{4pt}}}
\toprule
\aphead & \textbf{ARKitScenes} & \textbf{ScanNet++} & \textbf{Total} \\
\midrule
Scenes        & 41  & 31  & 72 \\
Episodes      & 103 & 131 & 234 \\
Questions     & 425 & 569 & 994 \\
Unanswerable  & 25 (5.9\%) & 39 (6.9\%) & 64 (6.4\%) \\
\bottomrule
\end{tabularx}
\end{table}

The main results use the 227 episodes whose oracle route fits within the step budget (Appendix~\ref{app:overbudget}). They contain 962 questions, of which 899 are answerable and 63 unanswerable, together with 137 of the linked constraints.

\subsection{Human Verification}
\label{app:qc}

A stratified sample of episodes from each source was verified by three expert annotators in two stages. The annotator first answered each question from the oracle views alone, without annotations and with the option to declare it undeterminable, and then compared each answer with the ground truth shown together with object boxes and a top-down map. A question passes if its ground truth is judged correct and its object labels are judged right. The pass rate was 96.7\% and 100\% in the two final rounds on ARKitScenes and 95.0\% on ScanNet++, rising to 98.1\% after the objects flagged as mislabeled were excluded from the benchmark. Questions flagged as ambiguous stayed below 5\%. An earlier round led us to drop size-based references such as ``the larger table'', which annotators could not resolve reliably. The first-stage answers form the human reference in Table~\ref{tab:main}.
\section{Implementation Setup}
\label{app:hyper}

\subsection{Models and Inference}
\label{app:models}

\paragraph{Interface.}
Proprietary models are queried through their official APIs and open-weight models through vLLM. All models receive the same prompts (Appendix~\ref{app:prompts}) and the same images at $640{\times}480$; ScanNet++ frames are downsampled to this size. Replies must be a single JSON object. We use the structured-output mode of each API where available and the JSON mode otherwise, without changing the prompt. Proprietary models run at the high reasoning effort of their provider. Since the two providers define this level independently, it is matched in name only. The output budget, which includes reasoning tokens, is set to 32{,}768 tokens per reply for each of the six proprietary models.

% Table~\ref{tab:truncation} is defined here, one page ahead of Sec.~\ref{app:truncation}, so that it lands at the top of the page that discusses it.
\begin{table*}[t]
\centering
\caption{\textbf{Truncation rate (\%) by model and setting.} Single-turn settings are counted per episode and Active per call. Ctx.: share of Active episodes ended early by context overflow. Rates above 15\% are in bold on a shaded background and correspond to the cells marked $^{\ddagger}$ in Table~\ref{tab:main}; zero rates are printed in grey.}
\label{tab:truncation}
\apsetup\setlength{\tabcolsep}{4pt}
\begin{tabularx}{0.86\textwidth}{@{\hspace{4pt}}l YYYYYYY >{\columncolor{asActive}}Y >{\columncolor{asActive}}Y@{\hspace{4pt}}}
\toprule
\aphead & \multicolumn{7}{c}{\textbf{Single-turn settings}} & \multicolumn{2}{c}{\textbf{Active}} \\
\aphead \textbf{Model} & Blind & P-8 & P-20 & P-40 & RW-40 & O-Min & O-Route & Trunc. & Ctx. \\
\midrule
\apgroup{10}{Proprietary models}
\mlogo{}All six models & \apzero & \apzero & \apzero & \apzero & \apzero & \apzero & \apzero & \apzero & \apzero \\
\apgroup{10}{Open-weight general models}
\mlogo{qwen}Qwen3.5-4B & 5.1 & \apflag{24.8} & \apflag{34.2} & \apflag{41.9} & \apflag{35.5} & \apflag{20.5} & 12.0 & \apflag{28.8} & \apzero \\
\mlogo{qwen}Qwen3.5-9B & \apzero & \apflag{42.3} & \apflag{44.0} & \apflag{39.7} & \apflag{62.8} & \apflag{26.1} & \apflag{33.3} & \apflag{18.6} & \apzero \\
\mlogo{qwen}Qwen3.6-27B & \apzero & 2.1 & 1.7 & 2.6 & 6.0 & 1.3 & 0.9 & 0.6 & \apzero \\
\mlogo{qwen}Qwen3.6-35B-A3B & \apzero & 3.0 & 4.7 & 3.0 & 5.1 & 0.9 & 3.4 & 0.9 & \apzero \\
\mlogo{qwen}Qwen3.8-27B & 2.1 & \apzero & \apzero & \apzero & \apzero & \apzero & 1.3 & 0.5 & \apzero \\
\mlogo{gemma}Gemma-4-31B-it & \apzero & \apzero & 0.4 & 0.4 & \apzero & \apzero & \apzero & 3.2 & \apflag{29.5} \\
\mlogo{internvl}InternVL3.5-8B & \apzero & \apzero & \apzero & \apzero & \apzero & \apzero & \apzero & \apzero & \apzero \\
\mlogo{llava}LLaVA-OneVision-7B & \apzero & \apzero & \apzero & \apzero & \apzero & \apzero & 0.4 & 0.1 & 0.4 \\
\mlogo{glm}GLM-4.6V-Flash & \apzero & \apzero & \apzero & \apzero & \apzero & \apzero & \apzero & \apzero & \apzero \\
\mlogo{kimi}Kimi-VL-A3B-Thinking & 3.5 & \apzero & 0.9 & \apzero & \apzero & 1.3 & \apzero & 0.6 & \apzero \\
\apgroup{10}{Spatial-specialist models}
\mlogo{sensenova}SenseNova-SI-1.1-2B & \apzero & \apzero & \apzero & \apzero & \apzero & \apzero & \apzero & \apzero & \apzero \\
\mlogo{sensenova}SenseNova-SI-1.2-8B & \apzero & \apzero & \apzero & \apzero & \apzero & \apzero & \apzero & \apzero & \apzero \\
\mlogo{sensenova}SenseNova-SI-1.3-8B & \apzero & \apzero & \apzero & \apzero & \apzero & \apzero & \apzero & \apzero & \apzero \\
\mlogo{vst}VST-7B-RL & \apzero & \apzero & \apzero & \apzero & \apzero & \apzero & \apzero & \apzero & \apzero \\
\mlogo{cambrian}Cambrian-S-7B & \apzero & \apzero & \apzero & \apzero & \apzero & \apzero & \apzero & 0.8 & \apzero \\
\mlogo{robobrain}RoboBrain2.0-7B & \apzero & \apzero & \apzero & \apzero & \apzero & \apzero & \apzero & \apzero & \apzero \\
\mlogo{cosmos}Cosmos-Reason1-7B & 2.2 & 3.5 & 1.8 & 0.4 & 5.3 & 1.8 & 11.5 & \apzero & \apzero \\
\bottomrule
\end{tabularx}
\end{table*}

\paragraph{Active episodes.}
An episode starts with the start image and all its questions. Each of the six moves costs one step, including a move that fails because no view exists in that direction, and so does a reply that cannot be parsed or names an illegal action. Three consecutive invalid replies end the episode with its questions unanswered. Answering is free, and a model that has used all 40 steps must answer. After every move the model receives the new image, the displacement and rotation of that step, and its cumulative pose relative to the start, computed from the ground-truth camera poses. The context keeps the start image and the eight most recent images, while the full record of actions, odometry, failed moves, and step count is kept as text. Image identifiers are randomized per episode, and frame indices, object boxes, visibility labels, and oracle information are hidden.

\paragraph{Reference settings.}
Blind gives the questions only and does not accept \texttt{cannot\_determine}, so it is scored on answerable questions. Passive-$k$ gives the start image and $k$ views sampled uniformly from the scene, shuffled and without poses. RW-40 takes 40 steps, each chosen uniformly among the moves that succeed, and gives all distinct views in the order visited with the actions and cumulative poses. O-Min gives the start image and the minimal set of evidence views, shuffled and without poses. O-Route gives every view along the shortest route that reveals the evidence of all questions in the episode, in order with actions and poses, and is thus the closest counterpart of the active setting.

\paragraph{Scoring and intervals.}
Multiple-choice questions are scored by accuracy and absolute distance by mean relative accuracy, $\frac{1}{10}\sum_{\theta}\mathbb{1}\big[\,|\hat{d}-d|/d < 1-\theta\,\big]$ with $\theta\in\{0.50,0.55,\dots,0.95\}$. All intervals are 95\% bootstrap intervals over scenes with 2{,}000 resamples. Results are from a single run, except for Claude Haiku, whose Active and RW-40 scores are each averaged over three runs.

\subsection{Baselines}
\label{app:baselines}

\paragraph{Random guessing.}
For multiple-choice questions the expected score is the reciprocal of the number of options. For absolute distance it is the expected mean relative accuracy of a guess drawn uniformly from the range of ground-truth distances, 0.52 to 5.46\,m. A random guesser never abstains, so it scores zero on unanswerable questions, which lowers its score from 28.0 on answerable questions to 26.2 overall.

\paragraph{Human reference.}
The human reference comes from the first stage of human verification (Appendix~\ref{app:qc}). Expert annotators answered a subset of 196 questions, 11 of them unanswerable, from the start image and the oracle views, without poses or annotations and with the option to abstain. This matches the O-Min setting, on which GPT-6.1 Sol scores 83.1 on the same subset against 82.5 for the annotator. The reference shows that the questions are solvable from the right evidence. It was not collected through active exploration and therefore should not be read as a human upper bound for the active exploration setting itself.

\subsection{Truncation and Context Overflow}
\label{app:truncation}

A response is truncated when it degenerates into repetition, exhausts the output budget before finishing its reasoning, or times out. It is then scored as unanswered. An Active episode may also end early when the accumulated context exceeds the limit of the model. Table~\ref{tab:truncation} reports both rates. When over 15\% of the responses of a model in a setting are affected, the corresponding cell in Table~\ref{tab:main} is marked $^{\ddagger}$, excluded from the ranking, and not used to compute gaps. This applies to Qwen3.5-4B and Qwen3.5-9B in most settings and the Active setting of Gemma-4, where 29.5\% of the episodes overflow the context. We keep these models for completeness but draw no conclusion from marked cells.

\subsection{Reply Parsing}
\label{app:parsing}

A reply is valid when it contains one JSON object with a legal action, or with answers to the questions. Two lenient rules recover replies whose content is clear but whose format deviates, and they are applied identically to all models and settings. First, when the final message is empty and the reasoning field consists of a well-formed reply, that reply is used. Second, when the JSON object omits the action field but contains an answers object, it is treated as an answer. No other repair is attempted, and a reply that names an illegal action or contains no answers remains invalid. The second rule matters mainly for Kimi-VL, which omits the action field in 602 single-turn replies; without it, its Blind score would be 6.5 instead of 33.1. For the other models the rule affects at most seven replies in any setting and never changes a reported score by more than a single point.

\subsection{Episodes beyond the Step Budget}
\label{app:overbudget}

In seven of the 234 episodes, all from ScanNet++, the shortest route that reveals all evidence is longer than 40 steps. No policy can see everything within the budget in these episodes, and the route that the oracle follows is itself longer than what the active agent is allowed. They were run under every setting but are excluded from the main results, which cover the remaining 227 episodes and 962 questions. Table~\ref{tab:overbudget} reports the scores on the seven episodes alone.

With only 32 questions, the individual numbers are noisy, but the pattern matches the main results in a sharper form. Along the oracle route the proprietary models still score between 56 and 94, so the questions remain answerable from the right views. Yet their Active scores fall well below those in Table~\ref{tab:main}, and the gap to the oracle route widens to between 30 and 70 points for all, as expected when the evidence cannot be reached within the budget.

\begin{table}[!t]
\centering
\caption{\textbf{Scores on the seven episodes beyond the step budget (\%).} These episodes contain 32 questions and are excluded from Table~\ref{tab:main}. $^{\dagger}$: the input of one episode exceeds the context length of the model. $^{\ast}$: scored before the lenient parsing rules of Appendix~\ref{app:parsing} were applied.}
\label{tab:overbudget}
\apsetup\setlength{\tabcolsep}{2.4pt}
\begin{tabularx}{\columnwidth}{@{\hspace{3pt}}X cccc >{\columncolor{asActive}}c@{\hspace{3pt}}}
\toprule
\aphead \textbf{Model} & Blind & P-40 & RW-40 & O-Route & Active \\
\midrule
\apgroup{6}{Proprietary models}
\mlogo{claude}Claude Opus 5.5 & 37.7 & 57.2 & 48.1 & 90.0 & 58.8 \\
\mlogo{claude}Claude Sonnet 5.5 & 24.5 & 47.8 & 34.4 & 86.6 & 37.2 \\
\mlogo{claude}Claude Haiku 5.5 & 46.1 & 37.5 & 34.1 & 84.7 & 30.9 \\
\mlogo{openai}GPT-6.1 Sol & 51.6 & 62.2 & 12.5 & 93.4 & 38.4 \\
\mlogo{openai}GPT-6.1 Terra & 37.1 & 58.8 & 15.6 & 93.8 & 25.0 \\
\mlogo{openai}GPT-6 Luna & 22.9 & 48.8 & 25.0 & 56.2 & 26.9 \\
\apgroup{6}{Open-weight general models}
\mlogo{qwen}Qwen3.5-4B & 36.1 & 23.4 & 21.9 & 46.9 & 15.6 \\
\mlogo{qwen}Qwen3.5-9B & 36.1 & 34.7 & 15.6 & 18.8 & 21.9 \\
\mlogo{qwen}Qwen3.6-27B & 34.8 & 47.2 & 32.2 & 62.8 & 15.6 \\
\mlogo{qwen}Qwen3.6-35B-A3B & 31.3 & 33.4 & 28.1 & 44.4 & 23.4 \\
\mlogo{qwen}Qwen3.8-27B & 39.0 & 56.6 & 25.0 & 56.2 & 9.4 \\
\mlogo{gemma}Gemma-4-31B-it & 39.0 & 50.3 & 15.6 & 39.4 & 15.6 \\
\mlogo{internvl}InternVL3.5-8B & 21.3 & 33.1 & 27.8 & 30.0 & 21.9 \\
\mlogo{llava}LLaVA-OneVision-7B & 21.3 & 20.3 & 13.1 & \na\rlap{$^{\dagger}$} & 27.8 \\
\mlogo{glm}GLM-4.6V-Flash & 39.0 & 40.3 & 22.5 & 48.8 & 12.5 \\
\mlogo{kimi}Kimi-VL-A3B-Thinking\rlap{$^{\ast}$} & 14.2 & 8.4 & 21.9 & 27.2 & 8.4 \\
\apgroup{6}{Spatial-specialist models}
\mlogo{sensenova}SenseNova-SI-1.1-2B & 32.3 & 24.4 & 21.2 & 22.5 & 27.5 \\
\mlogo{sensenova}SenseNova-SI-1.2-8B & 29.7 & 28.4 & 19.7 & 16.6 & 15.6 \\
\mlogo{sensenova}SenseNova-SI-1.3-8B & 30.0 & 15.0 & 21.9 & 21.2 & 14.4 \\
\mlogo{vst}VST-7B-RL & 24.5 & 30.0 & 28.7 & 35.0 & 10.6 \\
\mlogo{cambrian}Cambrian-S-7B & 29.0 & 6.2 & 9.1 & 6.2 & 16.2 \\
\mlogo{robobrain}RoboBrain2.0-7B & 34.8 & 25.9 & 27.2 & 29.1 & 15.6 \\
\mlogo{cosmos}Cosmos-Reason1-7B & 21.3 & 21.9 & 19.1 & 13.1 & 26.2 \\
\bottomrule
\end{tabularx}
\end{table}

\section{Additional Experimental Results}
\label{app:results}

\subsection{Robustness Checks}
\label{sec:ablation}

We ran two more experiments on GPT-6.1 Terra and Claude Haiku, representing the two exploration styles in Sec.~\ref{sec:behavior}: using the whole budget and stopping early (Table~\ref{tab:robust}).

\begin{table}[!htb]
\centering
\caption{\textbf{Robustness checks (\%).} \textit{Start view}: answer rate (AR) on answerable questions when a model sees only the start image and may abstain, next to its Blind score, where abstention is not allowed. \textit{Active score}: reasoning effort lowered from high to medium (Med.) on the same run. \textit{Steps}: median number of steps before answering. $^{\ast}$: 95\% interval excludes zero.}
\label{tab:robust}
\apsetup\setlength{\tabcolsep}{2.5pt}
\begin{tabularx}{\columnwidth}{@{\hspace{3pt}}l YY >{\columncolor{asActive}}Y>{\columncolor{asActive}}Y>{\columncolor{asActive}}Y YY@{\hspace{3pt}}}
\toprule
\aphead & \multicolumn{2}{c}{\textbf{Start view}} & \multicolumn{3}{c}{\textbf{Active score}} & \multicolumn{2}{c}{\textbf{Steps}} \\
\aphead \textbf{Model} & AR & Blind & High & Med. & $\Delta$ & High & Med. \\
\midrule
\mlogo{openai}GPT-6.1 Terra    & 1.4 & 44.2 & 55.1 & 56.7 & \gain{1.7}        & 40 & 39 \\
\mlogo{claude}Claude Haiku 5.5 & 6.8 & 39.0 & 42.7 & 37.7 & \gain{-5.0}\rlap{$^{\ast}$} & 13 & 9 \\
\bottomrule
\end{tabularx}
\end{table}

% Table~\ref{tab:persource} is defined here, ahead of Sec.~\ref{app:persource}, so that it lands at the top of the page that discusses it.
\begin{table*}[t]
\centering
\caption{\textbf{Additional reference settings and results by data source (\%).} P-8 and P-20: Passive-8 and Passive-20. O-Min: start image and minimal evidence views, without order or poses. ARKit: ARKitScenes; SN++: ScanNet++. $^{\ddagger}$: as in Table~\ref{tab:main}.}
\label{tab:persource}
\apsetup\renewcommand{\arraystretch}{1.0}\setlength{\tabcolsep}{4pt}
\begin{tabularx}{0.86\textwidth}{@{\hspace{4pt}}l YYY >{\columncolor{asActive}}Y>{\columncolor{asActive}}Y YY YY@{\hspace{4pt}}}
\toprule
\aphead & \multicolumn{3}{c}{\textbf{Reference settings (merged)}} & \multicolumn{2}{c}{\textbf{Active}} & \multicolumn{2}{c}{\textbf{RW-40}} & \multicolumn{2}{c}{\textbf{O-Route}} \\
\aphead \textbf{Model} & P-8 & P-20 & O-Min & ARKit & SN++ & ARKit & SN++ & ARKit & SN++ \\
\midrule
\apgroup{10}{Proprietary models}
\mlogo{claude}Claude Opus 5.5 & 62.2 & 70.2 & 75.2 & 70.6 & 76.7 & 73.2 & 67.7 & 86.7 & 84.8 \\
\mlogo{claude}Claude Sonnet 5.5 & 42.9 & 52.8 & 63.3 & 59.2 & 62.7 & 58.9 & 59.5 & 82.3 & 79.7 \\
\mlogo{claude}Claude Haiku 5.5 & 42.8 & 50.0 & 57.1 & 40.5 & 49.0 & 51.2 & 59.0 & 73.9 & 76.4 \\
\mlogo{openai}GPT-6.1 Sol & 53.8 & 73.8 & 83.1 & 65.9 & 71.0 & 56.3 & 66.2 & 89.8 & 89.6 \\
\mlogo{openai}GPT-6.1 Terra & 47.8 & 71.2 & 80.3 & 48.7 & 60.1 & 54.2 & 67.0 & 85.6 & 88.0 \\
\mlogo{openai}GPT-6 Luna & 52.4 & 56.5 & 62.9 & 35.8 & 43.9 & 51.8 & 55.8 & 63.8 & 70.4 \\
\apgroup{10}{Open-weight general models}
\mlogo{qwen}Qwen3.5-4B & 32.1\tr & 31.7\tr & 37.2\tr & 12.9\tr & 20.5\tr & 24.1\tr & 29.8\tr & 37.8 & 42.3 \\
\mlogo{qwen}Qwen3.5-9B & 27.1\tr & 29.7\tr & 38.9\tr & 20.6\tr & 25.3\tr & 14.4\tr & 16.3\tr & 34.1\tr & 33.7\tr \\
\mlogo{qwen}Qwen3.6-27B & 44.2 & 51.8 & 56.4 & 36.4 & 43.8 & 43.1 & 46.4 & 60.4 & 60.1 \\
\mlogo{qwen}Qwen3.6-35B-A3B & 44.9 & 52.6 & 58.7 & 35.2 & 41.7 & 46.2 & 43.2 & 52.5 & 52.8 \\
\mlogo{qwen}Qwen3.8-27B & 49.4 & 58.9 & 67.0 & 19.4 & 40.6 & 55.9 & 57.6 & 73.2 & 73.6 \\
\mlogo{gemma}Gemma-4-31B-it & 42.7 & 50.4 & 55.5 & 27.7\tr & 37.9\tr & 45.4 & 48.3 & 56.4 & 58.8 \\
\mlogo{internvl}InternVL3.5-8B & 33.6 & 32.4 & 34.5 & 23.9 & 28.6 & 31.8 & 34.8 & 34.9 & 35.6 \\
\mlogo{llava}LLaVA-OneVision-7B & 31.4 & 30.1 & 31.5 & 23.7 & 20.1 & 26.9 & 22.2 & 32.7 & 25.0 \\
\mlogo{glm}GLM-4.6V-Flash & 40.7 & 46.5 & 47.8 & 25.1 & 25.0 & 34.0 & 38.9 & 39.0 & 45.6 \\
\mlogo{kimi}Kimi-VL-A3B-Thinking & 33.0 & 32.5 & 40.2 & 27.7 & 26.6 & 33.5 & 32.9 & 37.0 & 37.3 \\
\apgroup{10}{Spatial-specialist models}
\mlogo{sensenova}SenseNova-SI-1.1-2B & 33.2 & 33.3 & 34.1 & 33.6 & 29.3 & 34.1 & 28.8 & 34.0 & 30.2 \\
\mlogo{sensenova}SenseNova-SI-1.2-8B & 34.1 & 35.1 & 35.1 & 26.2 & 19.9 & 33.0 & 33.4 & 35.5 & 33.0 \\
\mlogo{sensenova}SenseNova-SI-1.3-8B & 34.4 & 34.9 & 34.8 & 14.8 & 14.1 & 34.8 & 34.4 & 35.8 & 34.6 \\
\mlogo{vst}VST-7B-RL & 34.9 & 34.9 & 34.4 & 34.4 & 28.4 & 34.8 & 30.5 & 36.1 & 32.7 \\
\mlogo{cambrian}Cambrian-S-7B & 21.3 & 17.2 & 21.7 & 14.1 & 15.1 & 16.2 & 13.6 & 18.8 & 15.8 \\
\mlogo{robobrain}RoboBrain2.0-7B & 36.6 & 37.4 & 36.2 & 24.5 & 27.4 & 35.4 & 35.5 & 35.3 & 34.6 \\
\mlogo{cosmos}Cosmos-Reason1-7B & 30.4 & 31.2 & 31.4 & 34.3 & 28.3 & 23.9 & 25.0 & 24.5 & 27.0 \\
\bottomrule
\end{tabularx}
\end{table*}

\myfinding{The start view alone does not support an answer}
Given only the start image and allowed to abstain, Terra answers 1.4\% of the answerable questions and Haiku 6.8\%. Their Blind scores of 44.2 and 39.0, obtained when abstention is not allowed, therefore reflect forced guessing and not information in the start view, which in turn confirms that the questions cannot be answered without active exploration.

\myfinding{The conclusions are stable across reasoning effort and runs}
Lowering the reasoning effort from high to medium leaves Terra unchanged and costs Haiku 5.0 points, mostly because it explores less, with its median stopping step falling from 13 to 9. The high-effort score of Haiku in Table~\ref{tab:robust} is the single run paired with the medium-effort run and thus differs from the three-run average of 45.2 in Table~\ref{tab:main}. Across the three runs, its Active score varies by about 2.5 points around the mean, which indicates the run-to-run variation behind the single-run results of the other models.

\subsection{Results by Data Source and Additional Reference Settings}
\label{app:persource}

Table~\ref{tab:persource} reports the three reference settings omitted from Table~\ref{tab:main} and splits the main scores by data source.

\paragraph{Reference settings.}
For the proprietary models and the larger open-weight models, scores rise steadily from Passive-8 to Passive-20 and Passive-40 as more of the scene is covered. O-Min, which shows only the minimal evidence views without order or poses, scores below O-Route for every proprietary model, by 6.6 points for GPT-6.1 Sol and 10.4 for Claude Opus. The ordered route with poses therefore carries information beyond the evidence views themselves, which is why we use O-Route as the upper reference for the active setting. The spatially specialized models stay within a few points of their Blind score in all of them.

% Table~\ref{tab:consistency} is defined here, slightly ahead of Sec.~\ref{app:consistency}, so that it lands in the column next to that section.
\begin{table}[!t]
\centering
\caption{\textbf{Violation rate of linked questions (\%).} Dist.: constraints between two distances and their comparison (84 in total). Persp.: constraints between mirrored perspective questions (53 in total). Rates are over the constraints whose questions were all answered.}
\label{tab:consistency}
\apsetup\renewcommand{\arraystretch}{1.0}\setlength{\tabcolsep}{3pt}
\begin{tabularx}{\columnwidth}{@{\hspace{3pt}}l >{\columncolor{asActive}}Y>{\columncolor{asActive}}Y YY@{\hspace{3pt}}}
\toprule
\aphead & \multicolumn{2}{c}{\textbf{Active}} & \multicolumn{2}{c}{\textbf{O-Route}} \\
\aphead \textbf{Model} & Dist. & Persp. & Dist. & Persp. \\
\midrule
\apgroup{5}{Proprietary models}
\mlogo{claude}Claude Opus 5.5 & 0.0 & 15.1 & 0.0 & 3.8 \\
\mlogo{claude}Claude Sonnet 5.5 & 1.3 & 23.5 & 0.0 & 5.7 \\
\mlogo{claude}Claude Haiku 5.5 & 2.9 & 36.7 & 1.2 & 26.4 \\
\mlogo{openai}GPT-6.1 Sol & 0.0 & 18.4 & 0.0 & 1.9 \\
\mlogo{openai}GPT-6.1 Terra & 0.0 & 10.3 & 0.0 & 13.2 \\
\mlogo{openai}GPT-6 Luna & 0.0 & 30.6 & 0.0 & 18.9 \\
\apgroup{5}{Open-weight general models}
\mlogo{qwen}Qwen3.5-4B & 5.9 & 93.9 & 2.8 & 80.4 \\
\mlogo{qwen}Qwen3.5-9B & 7.7 & 97.0 & 2.1 & 67.5 \\
\mlogo{qwen}Qwen3.6-27B & 1.7 & 91.1 & 0.0 & 55.8 \\
\mlogo{qwen}Qwen3.6-35B-A3B & 1.4 & 74.5 & 0.0 & 68.6 \\
\mlogo{qwen}Qwen3.8-27B & 0.0 & 33.3 & 0.0 & 23.1 \\
\mlogo{gemma}Gemma-4-31B-it & 0.0 & 51.4 & 0.0 & 28.3 \\
\mlogo{internvl}InternVL3.5-8B & 50.0 & 90.6 & 48.7 & 84.9 \\
\apgroup{5}{Spatial-specialist models}
\mlogo{sensenova}SenseNova-SI-1.2-8B & 49.1 & 86.8 & 36.1 & 90.6 \\
\mlogo{sensenova}SenseNova-SI-1.3-8B & 36.1 & 89.5 & 46.2 & 84.9 \\
\mlogo{vst}VST-7B-RL & 26.7 & 92.5 & 30.9 & 94.2 \\
\mlogo{cambrian}Cambrian-S-7B & 66.7 & 81.8 & 38.5 & 81.0 \\
\bottomrule
\end{tabularx}
\end{table}

\paragraph{Agreement between sources.}
Model rankings on ARKitScenes and ScanNet++ agree closely. Over the 23 models, Kendall's $\tau$ between the two rankings is 0.71 for Active, 0.79 for RW-40, 0.83 for Passive-40, and 0.90 for O-Route, and the five best models under Active are the same and in the same order on both sources. The agreement is lower for $\Delta_{\mathrm{RW}}$ ($\tau=0.66$), as expected for a difference of two scores. Most models score somewhat higher on ScanNet++ in the active setting, which we attribute to its denser keyframes and richer object annotations. The main conclusions of Sec.~\ref{sec:main} therefore hold on each source separately.

\subsection{Consistency of Linked Answers}
\label{app:consistency}

Linked questions test whether the answers of a model agree with one another. A distance constraint pairs two absolute distances with the comparison between them and is violated when the comparison contradicts the two estimates. A perspective constraint pairs two perspective questions with the standpoint and facing object exchanged, whose correct answers are mirror images, and is violated when the two answers are not. A constraint is evaluated only when all its questions are answered. Table~\ref{tab:consistency} reports the violation rates for each of the 17 models of the first evaluation round.

Three observations follow. First, distance constraints are almost never violated by the proprietary and larger open-weight models, so a model that estimates two distances compares them consistently. Second, perspective constraints are violated far more often, by 10 to 37\% for proprietary models in the active setting, and for the small open-weight and specialist models at rates near or above the 75\% expected by chance. Third, inconsistency is a symptom of error, not an independent failure: no constraint is violated when all its questions are answered correctly, and along the oracle route the overall violation rate of Claude Opus, GPT-6.1 Sol, and Claude Sonnet is below 2.5\%.

\begin{figure*}[t]
    \centering
    \includegraphics[width=\linewidth]{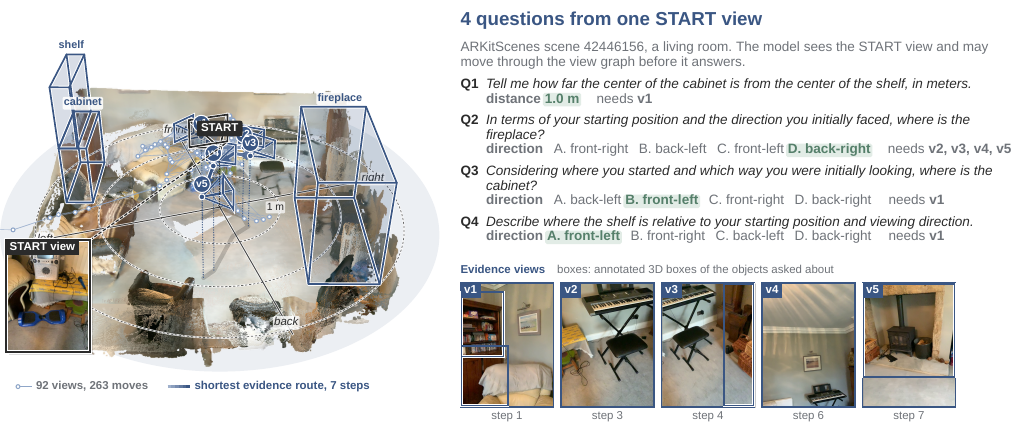}
    \vspace{-2em}
    \caption{\textbf{Benchmark Example: directions relative to the start.} An ARKitScenes living room with one distance and three direction questions. The cabinet and the shelf appear one step away, while the fireplace lies behind the START view and requires turning around.}
\label{fig:data1}
    \vspace{-1em}
\end{figure*}

\begin{figure*}[t]
    \centering
    \includegraphics[width=\linewidth]{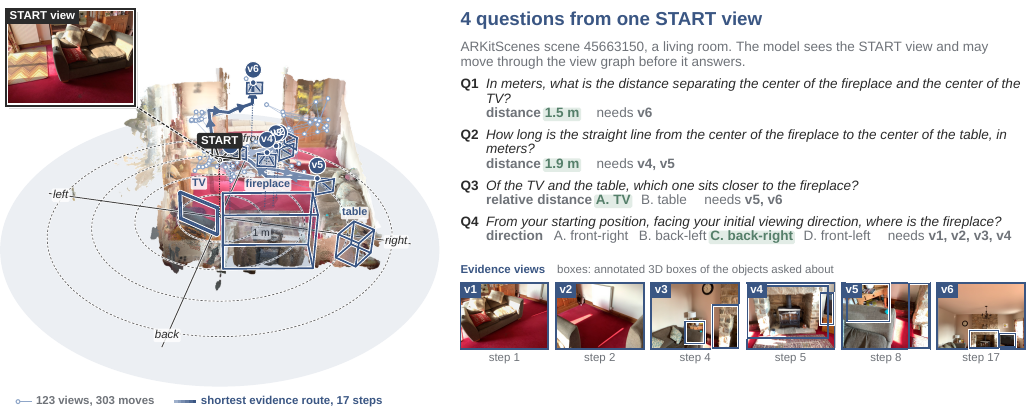}
    \vspace{-2em}
    \caption{\textbf{Benchmark Example: a linked constraint.} Two distance questions and their comparison share the fireplace, so the three answers must agree with each other. The last evidence view is reached only at step 17 of the shortest route.}
\label{fig:data2}
    \vspace{-1em}
\end{figure*}

\begin{figure*}[h]
    \centering
    \includegraphics[width=\linewidth]{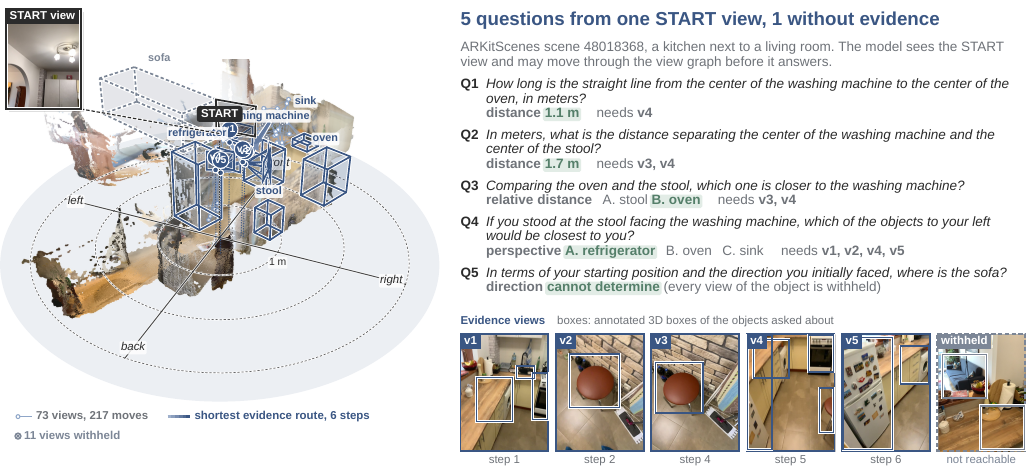}
    \vspace{-2em}
    \caption{\textbf{Benchmark Example: an unanswerable question.} A kitchen episode with five questions, including one that asks for a relation from an imagined viewpoint. All 11 views of the sofa are withheld, so the correct response to the last question is to abstain.}
\label{fig:data3}
    \vspace{-1em}
\end{figure*}

\begin{figure*}[h]
    \centering
    \includegraphics[width=\linewidth]{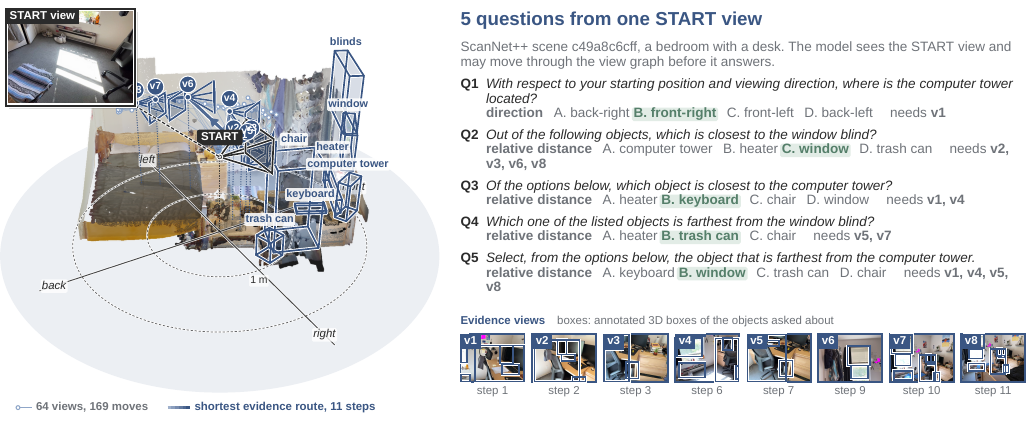}
    \vspace{-2em}
    \caption{\textbf{Benchmark Example: scattered evidence.} A ScanNet++ bedroom with one direction and four relative-distance questions. The evidence is spread over eight views along an 11-step route, far more than the recent views the model can keep in context.}
\label{fig:data4}
    \vspace{-1em}
\end{figure*}

\begin{figure*}[!t]
    \centering
    \includegraphics[width=\linewidth]{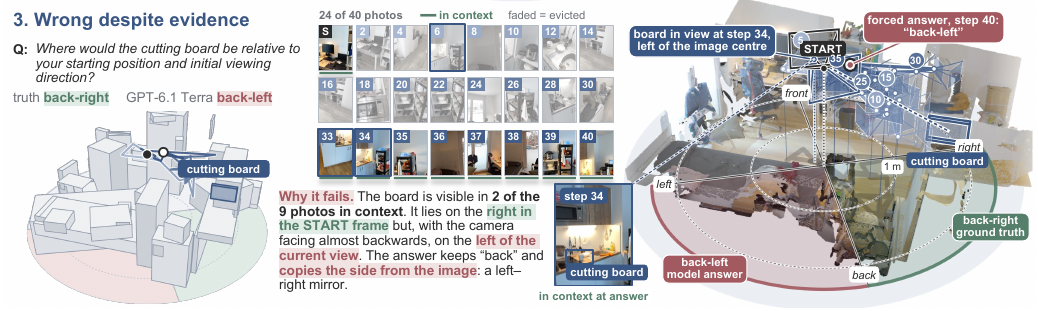}
    \vspace{-2em}
    \caption{\textbf{Representative failure: answering wrongly despite the evidence.} The cutting board is in view when GPT-6.1 Terra answers, but the answer takes the side from the current image and not from the start frame.}
    \label{fig:failures2}
    \vspace{-1em}
\end{figure*}

\begin{figure*}[!t]
    \centering
    \includegraphics[width=\linewidth]{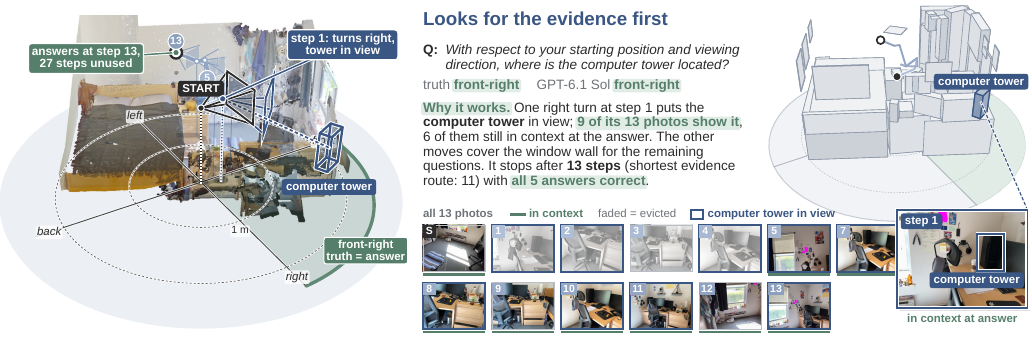}
    \vspace{-2em}
    \caption{\textbf{Representative success: looking for the evidence first.} GPT-6.1 Sol brings the target into view at step 1, keeps it in context, and stops after 13 steps with all five answers correct.}
    \label{fig:success1}
    \vspace{-1em}
\end{figure*}

\begin{figure*}[!t]
    \centering
    \includegraphics[width=\linewidth]{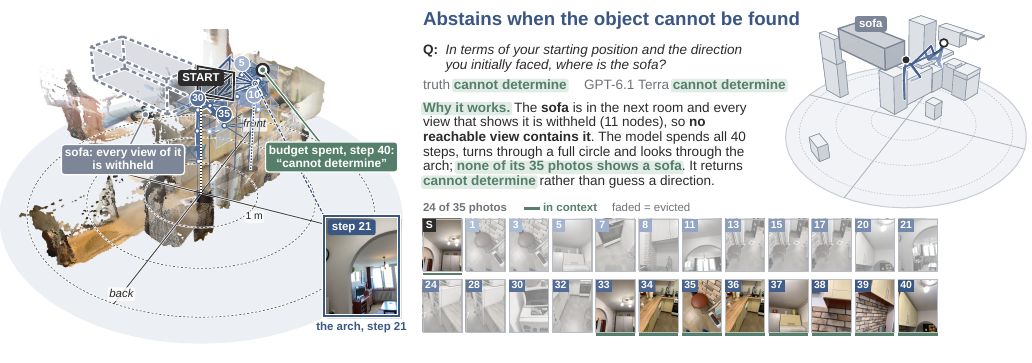}
    \vspace{-2em}
    \caption{\textbf{Representative success: abstaining when the evidence does not exist.} Every view of the sofa is removed. GPT-6.1 Terra searches the reachable space with all 40 steps and then answers \texttt{cannot\_determine}.}
    \label{fig:success2}
    \vspace{-1em}
\end{figure*}

\begin{figure*}[!t]
    \centering
    \includegraphics[width=\linewidth]{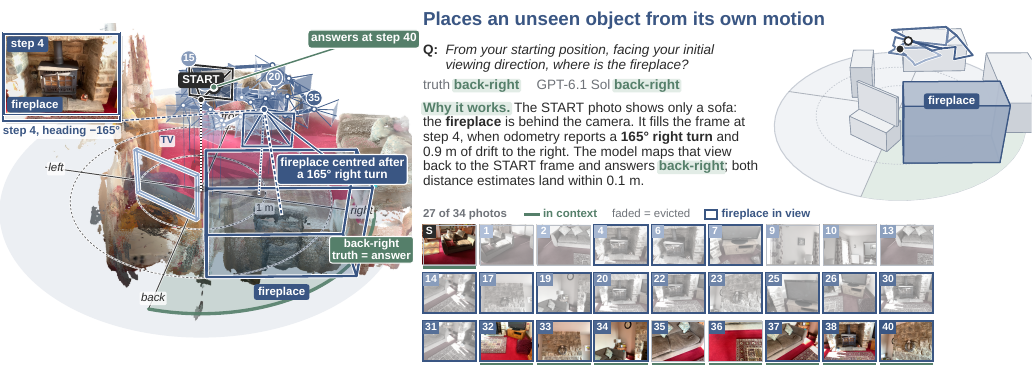}
    \vspace{-2em}
    \caption{\textbf{Representative success: locating an object from self-motion.} The fireplace is behind the start view. GPT-6.1 Sol sees it after a $165^{\circ}$ turn and places it correctly in the start frame.}
    \label{fig:success3}
    \vspace{-1em}
\end{figure*}

\begin{figure*}[h]
    \centering
    \includegraphics[width=\linewidth]{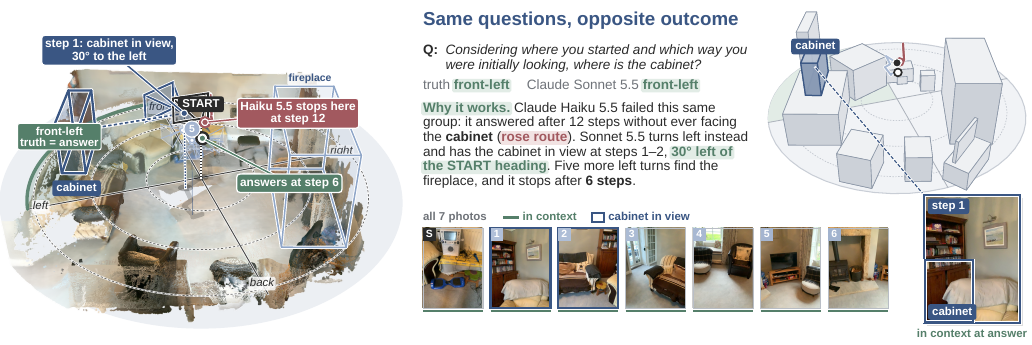}
    \vspace{-2em}
    \caption{\textbf{Representative success: the same episode, a different outcome.} On the episode where Claude Haiku stops without facing the cabinet (Fig.~\ref{fig:cases}), Claude Sonnet turns toward it at step 1 and answers correctly after six steps.}
    \label{fig:success4}
    \vspace{-1em}
\end{figure*}
\subsection{Cognitive-Map Probe}
\label{app:cogmap}

To test what a model retains about the scene when it answers, we probe GPT-6.1 Terra and Claude Haiku once more at the end of each episode, with the context unchanged. The model is asked, for every object mentioned in the questions, whether it knows where the object is and, if so, to give its position relative to the start pose. We compare the reported positions with the ground truth and group the objects by their state at answer time: seen and still among the images in context, seen but already evicted, and never seen. Table~\ref{tab:cogmap} reports the results for both models.

Both models locate objects that are still in context reliably, with a median error of about 0.4 to 0.5\,m. Once the images of an object have left the context, Terra reports a position for only 14.6\% of them and Haiku for 47.2\%, and the positions that are reported are far less accurate, with median errors of 1.4 and 2.1\,m. The textual record of actions and odometry, which remains in context, is thus not used to keep track of objects seen earlier. This supports the reading in Sec.~\ref{sec:bottleneck} that Terra declines questions about evicted evidence because it no longer knows where the objects are, not because it misremembers them. On the oracle route, with all evidence in context, both models locate nearly every object.

\begin{table}[!htb]
\centering
\caption{\textbf{Cognitive-map probe.} Located: share of objects for which the model reports a position (\%). Err.: median position error of the reported objects. Quad.: share of reported objects placed in the correct quadrant relative to the start pose (\%). The first three rows of each model are from the active setting; objects whose images have left the context are shaded.}
\label{tab:cogmap}
\apsetup\setlength{\tabcolsep}{3pt}
\begin{tabularx}{\columnwidth}{@{\hspace{3pt}}l YYYY@{\hspace{3pt}}}
\toprule
\aphead \textbf{Object state} & $\boldsymbol{n}$ & \textbf{Located} & \textbf{Err.\,(m)} & \textbf{Quad.} \\
\midrule
\apgroup{5}{\mlogo{openai}GPT-6.1 Terra}
\hspace{0.6em}Seen, in context & 736 & 91.4 & 0.42 & 83.1 \\
\rowcolor{apFlag}\hspace{0.6em}Seen, evicted    & 171 & 14.6 & 1.42 & 52.0 \\
\hspace{0.6em}Never seen       & 82  & 18.3 & 1.93 & 60.0 \\
\apsep
\hspace{0.6em}O-Route, in context & 989 & 97.7 & 0.28 & 89.6 \\
\apgroup{5}{\mlogo{claude}Claude Haiku 5.5}
\hspace{0.6em}Seen, in context & 633 & 90.8 & 0.50 & 77.6 \\
\rowcolor{apFlag}\hspace{0.6em}Seen, evicted    & 36  & 47.2 & 2.06 & 35.3 \\
\hspace{0.6em}Never seen       & 320 & 29.7 & 2.39 & 31.6 \\
\apsep
\hspace{0.6em}O-Route, in context & 989 & 98.8 & 0.41 & 82.3 \\
\bottomrule
\end{tabularx}
\end{table}

\section{Case Study}
\label{app:case_study}

This section complements the two failures in Fig.~\ref{fig:cases} with the third source of lost score and with four episodes in which exploration succeeds. Every case shows the scene from above with the route taken, the photographs collected along it, and the question with the ground truth and the answer of the model. Photographs that have already left the context by the time the model answers are shown faded.

\paragraph{Benchmark examples.}
Figures~\ref{fig:data1}--\ref{fig:data4} each show one episode of \textsc{ActSpace-Bench}. The left panel shows the reconstructed scene with its view graph, the START pose, and the shortest route that collects all evidence. The right panel lists the questions that share this START view, with their ground-truth answers and the evidence views each one needs. The bottom row shows these evidence views with the annotated 3D boxes of the queried objects, and the step at which the shortest route reaches them. The four episodes cover the main question types, a linked constraint (Fig.~\ref{fig:data2}), and an unanswerable question (Fig.~\ref{fig:data3}). They are also the episodes behind the model cases discussed below (Figs.~\ref{fig:success4}, \ref{fig:success3}, \ref{fig:success2}, and \ref{fig:success1}, respectively).

\paragraph{Answering wrongly despite the evidence.}
Fig.~\ref{fig:failures2} completes the three sources of loss in Sec.~\ref{sec:bottleneck}. GPT-6.1 Terra is asked where the cutting board lies relative to its start pose. The board is visible in two of the nine photographs in context when it answers, so the evidence is neither unfound nor evicted. In the start frame the board lies behind the agent and to its right. In the last view that shows it, however, the camera faces almost backwards, and the board appears on the left of the image. The answer keeps ``back'' but copies the side from the image, giving back-left for a back-right target. This is the single-axis error analyzed in Sec.~\ref{sec:errors}: the object is located, but the current view is not transformed back into the start frame, even though the heading relative to the start is reported with every single observation.

\paragraph{Looking for the evidence first.}
In Fig.~\ref{fig:success1}, GPT-6.1 Sol brings the queried object into view with its first action and keeps it there, so that six photographs of the object remain in context when it answers. Its other moves cover the part of the room needed for the remaining questions. It stops after 13 steps, close to the 11 steps of the shortest evidence route, and answers all five questions correctly. The episode shows the behavior that the benchmark is designed to reward: exploration directed at the evidence, and a stopping point set by the collected evidence and not by the budget.

\paragraph{Abstaining when the evidence does not exist.}
In Fig.~\ref{fig:success2}, the queried sofa stands in an adjacent room, and every view that shows it has been removed from the scene. GPT-6.1 Terra uses all 40 steps, turns through a full circle, and looks through the archway toward the other room. None of its 35 photographs shows a sofa, and it answers \texttt{cannot\_determine} instead of guessing a direction. The abstention follows a search that covers the reachable space, which clearly distinguishes it from the indiscriminate abstention behavior discussed earlier in Sec.~\ref{sec:abstention}.

\paragraph{Locating an object from self-motion.}
In Fig.~\ref{fig:success3}, the start photograph shows only a sofa, and the queried fireplace is behind the camera. It fills the frame at step 4, by which time the odometry reports a turn of $165^{\circ}$ to the right and a drift of 0.9\,m. GPT-6.1 Sol maps this view back into the start frame and answers back-right, and its two distance estimates in the same episode fall within 0.1\,m of the truth. This is the transformation that fails in Fig.~\ref{fig:failures2}, and it explains why start-relative direction is the task on which the strong models benefit the most from their own active exploration.

\paragraph{The same episode, a different outcome.}
Fig.~\ref{fig:success4} revisits the first failure in Fig.~\ref{fig:cases}, in which Claude Haiku turned in place near the start, never faced the cabinet, and answered after 12 steps. Claude Sonnet turns left instead and has the cabinet in view after one step, $30^{\circ}$ left of the start heading. Five more left turns find the fireplace needed for another question, and it stops after six steps, answering correctly. Both stop early, so the difference lies in where they looked, not in how long they explored, as in Sec.~\ref{sec:behavior}, where purposeful search does not depend on exploration length.

% \section{Full Related Works}
\newpage
\onecolumn
% Reset the page height: if the previous page holds only figures, \onecolumn would otherwise
% keep the reduced column height of the last two-column page and break the first box far too early.
\makeatletter\global\@colht\textheight\global\@colroom\textheight\global\vsize\textheight\makeatother
\section{Prompt Design}
\label{app:prompts}

All models receive the same prompts, shown verbatim below. Each setting uses one system prompt and one user message that lists the questions and the photographs. Text in angle brackets is filled in at run time, and $\langle$image$\rangle$ marks where a photograph is inserted. The conventions and answer rules are shared by all settings and are printed once, in the prompt of the active setting. Questions are listed as \texttt{Q1 (multiple choice)} followed by the question and its lettered options, or as \texttt{Q1 (numeric, meters)} followed by the question; replies must refer to the questions by these same IDs.

\definecolor{boxblue}{HTML}{6FA8DC}
\definecolor{boxblueback}{HTML}{EEF6FC}
\tcbset{promptbox/.style={colframe=boxblue, colback=boxblueback, coltitle=white, breakable, fonttitle=\bfseries}}

\subsection{Active Setting}

\begin{tcolorbox}[promptbox, title={System Prompt: Active}]
\small
You are exploring an indoor scene in order to answer questions about it.\\

You start at a fixed viewpoint and see the photograph taken there. You can move between nearby viewpoints; after every move you see the photograph taken at your new viewpoint. All questions are given at the start and are answered together at the end.\\

\textbf{Actions}\\
- \texttt{"forward"}, \texttt{"backward"}, \texttt{"left"}, \texttt{"right"}: move to a nearby viewpoint in that direction, relative to the direction you are currently facing. A move covers about 0.3 m.\\
- \texttt{"turn\_left"}, \texttt{"turn\_right"}: turn to a nearby viewpoint that faces further to the left or right, by about 45 degrees (your position may also shift by about 0.3 m).\\
- Viewpoints are not on a regular grid, so the actual change differs from move to move; the odometry below reports it.\\
- \texttt{"answer"}: submit your answers to all questions. This ends the task.\\

\textbf{Budget}\\
- You have 40 steps. Every move action uses one step, including a move that fails because there is no viewpoint in that direction (``Cannot move''); your position then stays the same. Turns can fail in the same way as other moves.\\
- A reply that cannot be read as a single JSON object, or that does not name a valid action, is invalid: it also uses one step, and you do not move. After 3 invalid replies in a row the task ends, and all questions count as unanswered and wrong.\\
- Answering does not use a step. You may answer at any time. When no steps remain, you must answer.\\

\textbf{Odometry}\\
- After each successful move you are told how your position and heading changed, measured relative to the direction you were facing before the move: the forward/backward and left/right distance in meters (rounded to 0.1 m) and how far you turned left or right in degrees (rounded to 5 degrees). Components that round to zero are omitted.\\
- With every observation you are also given your pose relative to START, in fixed coordinates defined by your starting position and starting viewing direction, measured on the floor plane: ``forward'' is the distance along the starting viewing direction (negative = behind the starting position), ``left'' is the distance to the left of it (negative = to the right), both in meters (rounded to 0.1 m); ``heading'' is the angle from the starting viewing direction to your current viewing direction, positive = turned to the left, between $-180$ and 180 degrees (rounded to 5 degrees). At START the pose is forward 0.0 m, left 0.0 m, heading 0 degrees.\\

\textbf{Photographs}\\
- Each photograph has an ID such as photo\_123456. Returning to a viewpoint you have already visited shows the same ID.\\
- The START photograph is always shown. Of the other photographs, only the 8 most recent are shown as images; earlier ones remain listed by ID in the text record of your moves.\\

\textbf{Conventions}\\
- All questions refer to the same indoor scene. Each object name in a question refers to one specific object in the scene.\\
- Distances are measured between the centers of the two objects, in meters.\\
- Directions are judged on the floor plane (heights are ignored), for a viewer standing at the stated position and facing the stated direction: an object is in ``front'' if the direction to its center is less than 90 degrees from the facing direction and ``back'' if it is more than 90 degrees; ``left'' and ``right'' are the two sides of the facing direction. Standing at an object means standing at its center; facing an object means facing toward its center.\\
- Your starting position and starting viewing direction are the position of the camera that took the START photograph and the horizontal direction in which it was pointing.\\

\textbf{Answers}\\
- Multiple-choice questions: give the option letter, for example \texttt{"B"}.\\
- Numeric questions: give a number in meters, for example 2.4.\\
- If the information available to you does not allow you to determine the answer to a question, give \texttt{"cannot\_determine"} for that question.\\
- Answer every question, using the question IDs (Q1, Q2, ...).\\

\textbf{Reply format}\\
Reply with a single JSON object and nothing else.\\
- To move: \texttt{\{"action": "forward"\}} (or any other move action)\\
- To answer: \texttt{\{"action": "answer", "answers": \{"Q1": "B", "Q2": 2.4, "Q3": "cannot\_determine"\}\}}\\
The JSON object may also contain an optional \texttt{"notes"} field: a short string for your own observations (at most 300 characters; longer notes are cut off). Notes are not scored.
\end{tcolorbox}

\begin{tcolorbox}[promptbox, title={User Messages: Active}]
\small
\textbf{First message}\\
Questions\\
$\langle$questions$\rangle$\\

You are at your starting viewpoint, $\langle$photo ID$\rangle$ (START).\\
$\langle$image$\rangle$\\
Pose relative to START: forward 0.0 m, left 0.0 m, heading 0 degrees.\\
Steps used: 0 of 40; 40 remaining. Choose your next action.\\

\textbf{After a successful move}\\
Step $\langle t\rangle$: \texttt{"}$\langle$action$\rangle$\texttt{"}. $\langle$odometry of this step$\rangle$ You are now at $\langle$photo ID$\rangle$.\\
$\langle$image$\rangle$\\
Pose relative to START: forward $\langle x\rangle$ m, left $\langle y\rangle$ m, heading $\langle\psi\rangle$ degrees.\\
Steps used: $\langle t\rangle$ of 40; $\langle 40-t\rangle$ remaining. Choose your next action.\\

\textbf{After a failed move}\\
Step $\langle t\rangle$: \texttt{"}$\langle$action$\rangle$\texttt{"}. Cannot move: no viewpoint in that direction. You are still at $\langle$photo ID$\rangle$; no new photograph.\\

\textbf{After an invalid reply}\\
Step $\langle t\rangle$: your reply was not a valid action, so one step was used and you did not move. You are still at $\langle$photo ID$\rangle$. Reply with a single JSON object as described in the instructions.\\

\textbf{When the budget is exhausted}\\
No steps remain. Reply with the answer action now.\\

\textbf{An image that has left the context is replaced by}\\
{[}$\langle$photo ID$\rangle$: image no longer shown{]}
\end{tcolorbox}

\subsection{Reference Settings}

\begin{tcolorbox}[promptbox, title={System Prompt: RW-40 and O-Route}]
\small
You are answering questions about an indoor scene from photographs taken along a path through it.\\

The path starts at the viewpoint of the START photograph, which shows your starting position and the direction you are facing. The record lists every move along the path in order. Each viewpoint's photograph is given once, in the order in which the viewpoints were first visited; a later return to the same viewpoint refers back to its photograph by ID. You cannot move or request more photographs.\\

Each entry of the record gives the pose at that viewpoint relative to START, in fixed coordinates defined by the starting position and starting viewing direction, measured on the floor plane: ``forward'' is the distance along the starting viewing direction (negative = behind the starting position), ``left'' is the distance to the left of it (negative = to the right), both in meters (rounded to 0.1 m); ``heading'' is the angle from the starting viewing direction to the viewing direction of the photograph, positive = turned to the left, between $-180$ and 180 degrees (rounded to 5 degrees). Moves are named relative to the direction faced before the move: \texttt{"forward"}, \texttt{"backward"}, \texttt{"left"}, \texttt{"right"} move to a nearby viewpoint in that direction; \texttt{"turn\_left"}, \texttt{"turn\_right"} turn to a nearby viewpoint that faces further to the left or right.\\

$\langle$Conventions and Answers, as in the active setting$\rangle$\\

\textbf{Reply format}\\
Reply with a single JSON object and nothing else:\\
\texttt{\{"action": "answer", "answers": \{"Q1": "B", "Q2": 2.4, "Q3": "cannot\_determine"\}\}}\\
$\langle$Notes rule, as in the active setting$\rangle$
\end{tcolorbox}

\begin{tcolorbox}[promptbox, title={User Message: RW-40 and O-Route}]
\small
Questions\\
$\langle$questions$\rangle$\\

Path record ($\langle n\rangle$ moves, $\langle m\rangle$ photographs):\\
START $\langle$photo ID$\rangle$. Pose relative to START: forward 0.0 m, left 0.0 m, heading 0 degrees.\\
$\langle$image$\rangle$\\
Move 1: \texttt{"}$\langle$action$\rangle$\texttt{"} to $\langle$photo ID$\rangle$. Pose relative to START: forward $\langle x\rangle$ m, left $\langle y\rangle$ m, heading $\langle\psi\rangle$ degrees.\\
$\langle$image$\rangle$\\
Move 2: \texttt{"}$\langle$action$\rangle$\texttt{"} back to $\langle$photo ID$\rangle$ (photograph shown above). Pose relative to START: \dots\\
\dots\\
Answer all questions.
\end{tcolorbox}

\begin{tcolorbox}[promptbox, title={System Prompt: Passive-$k$ and O-Min}]
\small
You are answering questions about an indoor scene from photographs.\\

You are given $\langle n\rangle$ photographs taken at different viewpoints in the same scene. The first photograph, marked START, shows your starting position and the direction you are facing. The other photographs are not in any particular order. You cannot move or request more photographs.\\

$\langle$Conventions and Answers, as in the active setting$\rangle$\\

\textbf{Reply format}\\
Reply with a single JSON object and nothing else:\\
\texttt{\{"action": "answer", "answers": \{"Q1": "B", "Q2": 2.4, "Q3": "cannot\_determine"\}\}}\\
$\langle$Notes rule, as in the active setting$\rangle$
\end{tcolorbox}

\begin{tcolorbox}[promptbox, title={User Message: Passive-$k$ and O-Min}]
\small
Questions\\
$\langle$questions$\rangle$\\

Photographs ($\langle n\rangle$):\\
START $\langle$photo ID$\rangle$\\
$\langle$image$\rangle$\\
$\langle$photo ID$\rangle$\\
$\langle$image$\rangle$\\
\dots\\
Answer all questions.
\end{tcolorbox}

\begin{tcolorbox}[promptbox, title={System Prompt: Blind}]
\small
You are answering questions about an indoor scene. No photographs of the scene are available in this setting; answer using general knowledge of how indoor spaces are typically arranged.\\

$\langle$Conventions, as in the active setting$\rangle$\\

\textbf{Answers}\\
- Multiple-choice questions: give the option letter, for example \texttt{"B"}.\\
- Numeric questions: give a number in meters, for example 2.4.\\
- Give your best guess for every question; \texttt{"cannot\_determine"} is not accepted in this setting.\\
- Answer every question, using the question IDs (Q1, Q2, ...).\\

\textbf{Reply format}\\
Reply with a single JSON object and nothing else:\\
\texttt{\{"action": "answer", "answers": \{"Q1": "B", "Q2": 2.4\}\}}\\
$\langle$Notes rule, as in the active setting$\rangle$
\end{tcolorbox}

% WARNING: do not forget to delete the supplementary pages from your submission 
% \input{sec/X_suppl}

\end{document}